# Solving Set Constraint Satisfaction Problems using ROBDDs

**Peter Hawkins**                                     HAWKINSP@CS.MU.OZ.AU
**Vitaly Lagoon**                                     LAGOON@CS.MU.OZ.AU
*Department of Computer Science and Software Engineering*
*The University of Melbourne, VIC 3010, Australia*

**Peter J. Stuckey**                                  PJS@CS.MU.OZ.AU
*NICTA Victoria Laboratory, Department of Computer Science and Software Engineering*
*The University of Melbourne, VIC 3010, Australia*

## Abstract

In this paper we present a new approach to modeling finite set domain constraint problems using Reduced Ordered Binary Decision Diagrams (ROBDDs). We show that it is possible to construct an efficient *set domain* propagator which compactly represents many set domains and set constraints using ROBDDs. We demonstrate that the ROBDD-based approach provides unprecedented flexibility in modeling constraint satisfaction problems, leading to performance improvements. We also show that the ROBDD-based modeling approach can be extended to the modeling of integer and multiset constraint problems in a straightforward manner. Since domain propagation is not always practical, we also show how to incorporate less strict consistency notions into the ROBDD framework, such as set bounds, cardinality bounds and lexicographic bounds consistency. Finally, we present experimental results that demonstrate the ROBDD-based solver performs better than various more conventional constraint solvers on several standard set constraint problems.

## 1. Introduction

It is often natural to express a constraint satisfaction problem (CSP) using finite domain variables and relations between those variables, where the values for each variable are drawn from a finite universe of possible values. One of the most common methods of solving finite domain CSPs is through maintaining and updating a domain for each variable using the combination of a backtracking search and an incomplete local propagation algorithm. The local propagation algorithm attempts to enforce consistency on the values in the variable domains by removing values that cannot form part of a solution to the system of constraints. Various levels of consistency can be defined, each with associated costs and benefits. Most consistency algorithms are incomplete—that is, they are incapable of solving the problem by themselves, so they must be combined with a backtracking search procedure to produce a complete constraint solver.

When attempting to apply this general scheme to the task of solving constraint satisfaction problems over finite set variables we quickly run into practical problems. Since the universe of possible values for a set variable is usually very large, the naïve representation of the domain of a set variable as a set of sets is too unwieldy to solve realistic problems. For example, if a set variable can take on the value of any subset of the set $\{1, \ldots, N\}$, then its domain contains $2^N$ elements, which quickly becomes infeasible to represent as $N$ increases in magnitude. Accordingly, most set constraint solvers to date have used an approximation to the true domain of a set variable in order to avoid this combinatorial explosion.

The most common approximation to the true domain of set variable $v$ has been to maintain an upper bound $U$ and a lower bound $L$ of the domain under the subset partial ordering relation and to perform set *bounds* propagation on these bounds. That is, $L$ contains elements that must be in the set $v$, and $U$ is the complement of the set of elements that must not be in $v$. Conventionally, a fixed set of inference rules specific to each constraint are used to enforce consistency on these upper and lower bounds. This basic scheme was proposed by Puget (1992), and has been implemented by set solvers such as Conjunto (Gervet, 1997), the `fd_sets` and `ic_sets` libraries of ECL$^i$PS$^e$ (IC-PARC, 2003), Mozart (Müller, 2001), and ILOG Solver (ILOG, 2004).

Set bounds are a crude approximation to a set domain at best, and thus various refinements to the bounds propagation scheme have been proposed so as to more effectively capture the nature of a set domain. Azevedo (2002) demonstrated that maintaining and performing inferences upon the upper and lower bounds on the cardinality of a set domain leads to a significant performance improvement on a variety of problems. While earlier set solvers such as Conjunto also maintained cardinality bounds, only partial usage was made of this information. More recently, Sadler and Gervet (2004) showed that incorporating upper and lower bounds under a lexicographic ordering leads to significantly stronger propagation, albeit at the cost of a marked increase in propagation time, leading to only a marginal performance improvement overall. While both of these approaches provide more effective propagation than the simple set bounds scheme, they do not approach the effectiveness of a true set domain propagator, which ensures that every value in the domain of a set variable can be extended to a complete assignment of every variable in any given constraint.

Consequently, we would like to devise a representation for set domains and constraints on those domains that is tractable enough to permit domain propagation. The principal observation that permits the implementation of a set domain propagator is that a finite integer set $v$ can be represented by its characteristic function $\chi_v$:

$$\chi_v : \mathbb{Z} \to \{0, 1\} \text{ where } \chi_v(i) = 1 \text{ iff } i \in v$$

Accordingly we can use a set of Boolean variables $v_i$ to represent the set $v$, which correspond to the propositions $v_i \leftrightarrow i \in v$. We can describe set domains and set constraints in terms of these Boolean variables. Interestingly, set bounds propagation as described above is equivalent to performing domain propagation in a naïve way on this Boolean representation.

In this paper we investigate this Boolean modeling approach for modeling finite domain constraints, and in particular set constraints. We show that it is possible to represent the domains of set variables using Reduced Ordered Binary Decision Diagrams (ROBDDs), a data structure for representing and manipulating Boolean formulæ. Such representations are usually fairly compact, even if they correspond to very large domains. In addition, it is possible to represent the constraints themselves as ROBDDs, permitting us to produce efficient *set domain* constraint propagators solely using ROBDD operations.

The ROBDD-based representation allows us to easily conjoin constraints and existentially quantify variables, thus permitting us to remove intermediate variables and merge constraints for stronger propagation. The construction of global constraints becomes an almost trivial exercise, without the requirement to write laboriously new propagators for every new constraint that we would like to use.

We also demonstrate that only minor changes are needed to the operation of the set domain propagator in order to implement other, less strict notions of consistency. In particular, we show how to construct set bounds, cardinality bounds and lexicographic bounds propagators, and how to utilize these notions to produce a *split domain* solver that combines bounds and domain reasoning.

A key theme of this paper is the flexibility of the ROBDD-based modeling approach. We are not limited only to modeling set variables—for example ROBDDs can be used to model integer variables and integer constraints. While the ROBDD-based approach to modeling finite domain integer variables is in general not as efficient as many existing finite domain integer solvers, the ability to model integer constraints using ROBDDs is an essential building block which allows us to construct set constraints such as cardinality and weighted sum, as well as allowing us to represent multisets and multiset constraints.

Finally, we present experiments using a variety of standard set problems to demonstrate the advantages of our modeling approach.

Many of the ideas from this paper have been previously published in two previous works (Lagoon & Stuckey, 2004; Hawkins, Lagoon, & Stuckey, 2004). This paper contains a more complete exposition of those ideas, as well as important extensions of the work that has previously been presented. These include substantial improvements in the complexity of the propagation algorithm, as well as cardinality and lexicographic bounds solvers implemented using ROBDDs. In addition, we show how to model integer expressions, allowing us to implement a weighted-sum constraint, and to model multisets and multiset constraints. Using this, we present new experimental results for the Hamming Code and Balanced Academic Curriculum problems. Finally, we present results comparing the ROBDD-based solver against a solver with good cardinality reasoning (Mozart).

The remainder of this paper is structured as follows. Section 2 contains essential concepts and definitions necessary when discussing finite domain solvers and ROBDDs. Section 3 shows how to model set domains and set constraints using ROBDDs, and presents a basic outline of an ROBDD-based set constraint solver. Section 4 demonstrates how to improve the performance of the ROBDD-based set solver through the construction of global constraints, the removal of intermediate variables, and through symmetry-breaking approaches. Section 5 demonstrates how to model integer and multiset expressions, as well as how to implement a weighted-sum constraint for set and multiset variables. Section 6 shows how to construct a more efficient domain propagator, as well as set bounds, set cardinality, and lexicographic bounds propagators. Finally, in Section 7 we present experimental results comparing the ROBDD-based solver with more conventional set solvers on a variety of standard problems.

## 2. Preliminaries

In this section we define the concepts and notation necessary when discussing propagation-based constraint solvers. These definitions are largely identical to those presented by Lagoon and Stuckey (2004) and others. For simplicity we shall present all of our definitions in the case of finite set variables; the extensions to multiset and integer variables are trivial.

## 2.1 Lattices, Domains, and Valuations

Let $\mathcal{L}$ be the powerset lattice $\langle \mathcal{P}(\mathcal{U}), \subset \rangle$, where $\mathcal{P}(x)$ denotes the powerset of $x$ and the *universe* $\mathcal{U}$ is a finite subset of $\mathbb{Z}$. We say a subset $M \subseteq \mathcal{L}$ is *convex* if for any $a, c \in M$ the relation $a \subseteq b \subseteq c$ implies $b \in M$ for all $b \in \mathcal{L}$. The *interval* $[a, b]$ is the set $M = \{x \in \mathcal{L} \mid a \subseteq x \subseteq b\}$. Intervals are obviously convex. Given a subset $K \subseteq \mathcal{L}$, we define the convex closure of $K$:

$$conv(K) = \left[ \bigcap_{x \in K} x, \bigcup_{x \in K} x \right]$$

The convex closure operation satisfies the properties of extension ($x \subseteq conv(x)$), idempotence ($conv(x) = conv(conv(x))$), and monotonicity (if $x \subseteq y$ then $conv(x) \subseteq (y)$) (Gervet, 1997).

**Example 2.1.** The set $X = \{\{1\}, \{1, 3\}, \{1, 4\}, \{1, 3, 4\}\}$ is convex and equivalent to the interval $[\{1\}, \{1, 3, 4\}]$. Conversely, the set $Y = \{\{1\}, \{1, 3\}, \{1, 3, 4\}\}$ is not convex. However, the convex closure of $Y$ is precisely the interval $X$, i.e. $conv(Y) = X$.

Let $\mathcal{V}$ denote the fixed finite collection of all set variables. Each variable has a domain, which is a finite collection of possible values from $\mathcal{L}$ (which are themselves sets). More generally, we define a *domain $D$* to be a complete mapping from the set $\mathcal{V}$ to finite collections of finite sets of integers. When we speak of the *domain of a variable $v$*, we mean $D(v)$. A domain $D_1$ is said to be *stronger* than a domain $D_2$, written $D_1 \sqsubseteq D_2$, if $D_1(v) \subseteq D_2(v)$ for all $v \in \mathcal{V}$. Two domains are said to be equal, written $D_1 = D_2$, if $D_1(v) = D_2(v)$ for all $v \in \mathcal{V}$. We call a domain $D$ a *range domain* if $D(v)$ is an interval for all $v \in \mathcal{V}$. We extend the concept of convex closure to domains by defining $ran(D)$ to be the unique (range) domain such that $ran(D)(v) = conv(D(v))$ for all $v \in \mathcal{V}$.

A *valuation* is a function from $\mathcal{V}$ to $\mathcal{L}$, which we write using the mapping notation $\{v_1 \mapsto d_1, v_2 \mapsto d_2, \ldots, v_n \mapsto d_n\}$, where $v_i \in \mathcal{V}$ and $d_i \in \mathcal{L}$ for all $i \in \mathbb{N}$. Clearly a valuation can be extended to constraints involving the variables in the obvious way. We define *vars* to be the function that returns the set of variables that are involved in a constraint, expression or valuation. In an abuse of notation, a valuation $\theta$ is said to be an element of a domain $D$, written $\theta \in D$, if $\theta(v) \in D(v)$ for all $v \in vars(\theta)$.

We say a domain $D$ is a *singleton* or *valuation* domain if $|D(v)| = 1$ for all $v \in \mathcal{V}$. In this case $D$ corresponds to a single valuation $\theta$ where $D(v) = \{\theta(v)\}$ for all $v \in \mathcal{V}$.

## 2.2 Constraints, Propagators, and Propagation Solvers

A *constraint* is a restriction placed upon the allowable values for a collection of variables. We are interested in the following *primitive set constraints*, where $k$ is an integer, $d$ is a ground set value, and $u$, $v$ and $w$ are set variables: $k \in v$ (membership), $k \notin v$ (non-membership), $u = v$ (equality), $u = d$ (constant equality), $u \subseteq v$ (non-strict subset), $u = v \cup w$ (union), $u = v \cap w$ (intersection), $u = v \setminus w$ (set difference), $u = \overline{v}$ (complement), $u \neq v$ (disequality), $|u| = k$ (cardinality), $|u| \geq k$ (lower cardinality bound), and $|u| \leq k$ (upper cardinality bound). Later we shall introduce non-primitive set constraints which are formed by composing primitive set constraints.

We define the *solutions* of a constraint $c$ to be the set of valuations that make that constraint true, i.e. $solns(c) = \{\theta \mid vars(\theta) = vars(c) \text{ and } \models \theta(c)\}$.

**Example 2.2.** Suppose $v$ and $w$ are set variables, and $D$ is a domain with $D(v) = \{\{1\}, \{1, 3\}, \{2, 3\}\}$, and $D(w) = \{\{2\}, \{1, 2\}, \{1, 3\}\}$. Let $c$ be the constraint $v \subseteq w$. Then the solutions to $c$ in the domain $D$ are the valuations $\{v \mapsto \{1\}, w \mapsto \{1, 2\}\}$, $\{v \mapsto \{1\}, w \mapsto \{1, 3\}\}$, and $\{v \mapsto \{1, 3\}, w \mapsto \{1, 3\}\}$.

With every constraint we associate a *propagator* $f$, which is a monotonic decreasing function from domains to domains, so if $D_1 \sqsubseteq D_2$ then $f(D_1) \sqsubseteq f(D_2)$ and $f(D_1) \sqsubseteq D_1$. A propagator $f$ is said to be *correct* for a constraint $c$ if:

$$\{\theta \mid \theta \in D\} \cap solns(c) = \{\theta \mid \theta \in f(D)\} \cap solns(c)$$

Correctness is not a strong restriction, since the identity propagator is correct for all constraints $c$. We usually assume that all propagators are both correct and *checking*, that is, if $D$ is a singleton domain formed through propagation then its associated valuation $\theta$ makes $\theta(c)$ true.

We can use propagators to form the basis for a constraint solver. A *propagation solver* $solv(F, D)$ takes a set of propagators $F$ and a current domain $D$, and repeatedly applies propagators from $F$ to the current domain until a fixpoint is reached. In other words $solv(F, D)$ is the weakest domain $D' \sqsubseteq D$ which is a fixpoint (i.e. $f(D') = D'$) for all $f \in F$. This fixpoint is unique (Apt, 1999).

## 2.3 Local Consistency

The notion of local consistency is of importance when considering the solution of constraint satisfaction problems. We can define various levels of consistency of different strengths and levels of difficulty to enforce. We describe two here.

A domain $D$ is said to be *domain consistent* for a constraint $c$ if $D$ is the strongest domain which contains all solutions $\theta \in D$ of $c$; in other words there does not exist a domain $D' \sqsubseteq D$ such that $\theta \in D$ and $\theta \in solns(c)$ implies $\theta \in D'$. A set of propagators maintains domain consistency if $solv(F, D)$ is domain consistent for all constraints $c$.

**Definition 1.** *A* domain propagator *for a constraint $c$ is a function $dom(c)$ mapping domains to domains which satisfies the following identity:*

$$dom(c)(D)(v) = \begin{cases} \{\theta(v) \mid \theta \in D \ \wedge \ \theta \in solns(c)\} & \text{if } v \in vars(c) \\ D(v) & \text{otherwise} \end{cases}$$

**Lemma 2.1.** *A domain propagator $dom(c)$ for a constraint $c$ is correct, checking, monotonic, and idempotent.*

*Proof.* Straightforward from the definitions. □

**Example 2.3.** Consider the constraint $v \subseteq w$ and domain $D$ of Example 2.2. The domain propagation $D' = dom(c)(D)$ returns domain $D'$ where $D'(v) = \{\{1\}, \{1, 3\}\}$ and $D'(w) = \{\{1, 2\}, \{1, 3\}\}$. The missing values are those that did not take part in any solution.

Domain consistency may be difficult to achieve for set constraints, so instead we often need a weaker notion of consistency. The notion of set bounds consistency is commonly used. A domain $D$ is *bounds consistent* for a constraint $c$ if for every variable $v \in vars(c)$ the upper bound of $D(v)$ is the union of the values of $v$ in all solutions of $c$ in $D$, and the lower bound of $D(v)$ is the intersection of the values of $v$ in all solutions of $c$ in $D$. As above, a set of propagators $F$ is said to maintain set bounds consistency for a constraint $c$ if $solv(F, D)$ is bounds consistent for all domains $D$.

**Definition 2.** *A set bounds propagator for a constraint $c$ is a function $sb(c)$ mapping domains to domains satisfying the following identity:*

$$sb(c)(D)(v) = \begin{cases} conv(dom(c)(ran(D))(v)) & \text{if } v \in vars(c) \\ D(v) & \text{otherwise} \end{cases}$$

**Lemma 2.2.** *A set bounds propagator $sb(c)$ for a constraint $c$ is correct, checking, and idempotent. A propagator $sb(c)$ is also monotonic on range domains.*

*Proof.* All of these follow trivially from the properties of $dom(c)$ and from the extension and idempotence properties of $conv$ and $ran$. □

Clearly for any constraint $c$ and any $v \in vars(c)$ we have $dom(c)(D)(v) \subseteq sb(c)(D)(v)$ for all domain propagators $dom(c)$ and all set bounds propagators $sb(c)$.

### 2.4 Boolean Formulæ and Binary Decision Diagrams

We use Boolean formulæ extensively to model sets and set relations. In particular, we make use of the following Boolean operations: $\wedge$ (conjunction), $\vee$ (disjunction), $\neg$ (negation), $\rightarrow$ (implication), $\leftrightarrow$ (bi-directional implication), $\oplus$ (exclusive OR), and $\exists$ (existential quantification). We use the shorthand $\exists_V F$ for $\exists_{x_1} \cdots \exists_{x_n} F$ where $V = \{x_1, \ldots, x_n\}$, and we use $\overline{\exists}_V F$ to mean $\exists_{V'} F$ where $V' = vars(F) \setminus V$.

Binary Decision Trees (BDTs) are a well-known method of modeling Boolean functions on Boolean variables. A Binary Decision Tree is a complete binary tree, in which each internal node $n(v, t, f)$ is labelled with a Boolean variable $v$ and each leaf node is labelled with a truth value 0 (false) or 1 (true). Each internal node corresponds to an if-then-else test of the labelled variable, and has two outgoing arcs—the "false" arc (to BDT $f$) and the "true" arc (to BDT $t$). In order to evaluate the function represented by the binary tree, the tree is traversed from the root to a leaf. On reaching any internal node, the value of the Boolean variable whose label appears on the node is examined, and the corresponding arc is followed. The traversal stops upon reaching a leaf node, whereupon the value of the function is taken to be the label of that node.

A Binary Decision Diagram (BDD) is a variant of a Binary Decision Tree, formed by relaxing the tree structure requirement, instead representing functions as directed acyclic graphs. In a Binary Decision Diagram, a node is permitted to have multiple parents, as opposed to a tree structure which requires any node to have at most one parent. This effectively permits common portions of the tree to be shared between multiple branches, allowing a compact representation of many functions.

The primary disadvantage of both Binary Decision Trees and Binary Decision Diagrams is that the representation of a given function is not canonical (i.e. one function may have many representations). Two additional canonicity properties allow many operations on BDDs to be performed efficiently (Bryant, 1986, 1992). A BDD is said to be *reduced* if it contains no identical nodes (that is, there are no nodes with the same label and identical "then" and "else" arcs), and no redundant tests (there are no nodes with both "then" and "else" arcs to the same node). A BDD is said to be *ordered* if there is a total order $\prec$ of the variables, such that if there is an arc from a node labelled $v_1$ to a node labelled $v_2$ then $v_1 \prec v_2$. A *Reduced Ordered BDD* (ROBDD) is canonical function representation up to reordering, which permits an efficient implementation of many Boolean function operations. For more details the reader is referred to the work of Bryant (1992), or the introduction of Andersen (1998).

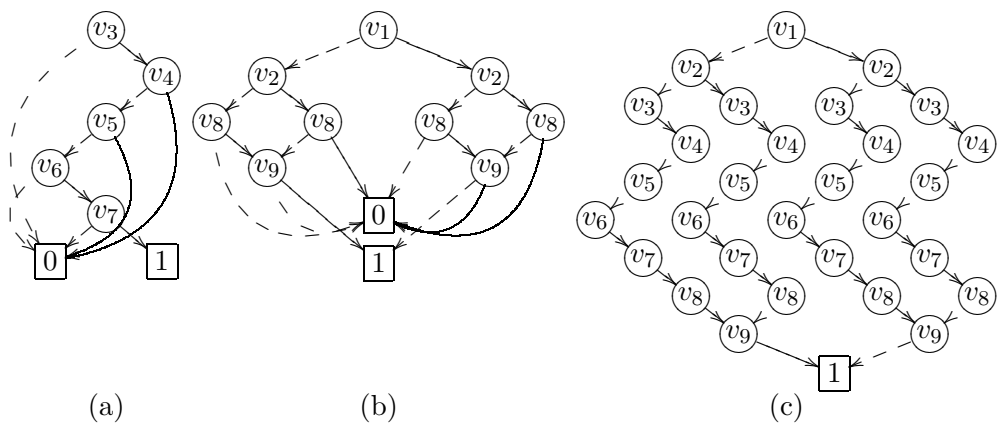

Figure 1: ROBDDs for (a) $L\overline{U} = v_3 \wedge \neg v_4 \wedge \neg v_5 \wedge v_6 \wedge v_7$ (b) $R = \neg(v_1 \leftrightarrow v_9) \wedge \neg(v_2 \leftrightarrow v_8)$ and (c) $L\overline{U} \wedge R$ (omitting the node 0 and arcs to it). Solid arcs are "then" arcs, dashed arcs are "else" arcs.

We define the size $|R|$ to be the number of non-leaf nodes of the ROBDD $R$, as well as defining $VAR(R)$ to be the set of ROBDD variables that appear as labels on the internal nodes of $R$. We shall be interested in *stick* ROBDDs, in which every internal node $n(v, t, f)$ has exactly one of $t$ or $f$ as the constant 0 node.

**Example 2.4.** Figure 1(a) gives an example of a stick ROBDD $L\overline{U}$ representing the formula $v_3 \wedge \neg v_4 \wedge \neg v_5 \wedge v_6 \wedge v_7$. $|L\overline{U}| = 5$ and $VAR(L\overline{U}) = \{v_3, v_4, v_5, v_6, v_7\}$. Figure 1(b) gives an example of a more complex ROBDD representing the formula $\neg(v_1 \leftrightarrow v_9) \wedge \neg(v_2 \leftrightarrow v_8)$. $|R| = 9$ and $VAR(R) = \{v_1, v_2, v_8, v_9\}$. One can verify that the valuation $\{v_1 \mapsto 1, v_2 \mapsto 0, v_8 \mapsto 1, v_9 \mapsto 0\}$ makes the formula true by following the path right, left, right, left from the root.

## 2.5 ROBDD Operations

There are efficient algorithms for many Boolean operations applied to ROBDDs. The complexity of the basic operations for constructing new ROBDDs is $O(|R_1||R_2|)$ for $R_1 \wedge R_2$,

$\mathsf{node}(v, t, f)$ **if** $(t = f)$ **return** $t$ **else return** $n(v, t, f)$

$\mathsf{and}(R_1, R_2)$
  **if** $(R_1 = 0 \textbf{ or } R_2 = 0)$ **return** $0$
  **if** $(R_1 = 1)$ **return** $R_2$
  **if** $(R_2 = 1)$ **return** $R_1$
  $n(v_1, t_1, f_1) := R_1$
  $n(v_2, t_2, f_2) := R_2$
  **if** $(v_1 \preceq v_2)$ **return** $\mathsf{node}(v_1, \mathsf{and}(t_1, R_2), \mathsf{and}(f_1, R_2))$
  **else if** $(v_2 \preceq v_1)$ **return** $\mathsf{node}(v_2, \mathsf{and}(t_2, R_1), \mathsf{and}(f_2, R_1))$
  **else return** $\mathsf{node}(v_1, \mathsf{and}(t_1, t_2), \mathsf{and}(f_1, f_2))$

$\mathsf{exists}(v, R)$
  **if** $(R = 0)$ **return** $0$
  **if** $(R = 1)$ **return** $1$
  $n(v_r, t, f) := R$
  **if** $(v_r \preceq v)$ **return** $\mathsf{node}(v_r, \mathsf{exists}(v, t), \mathsf{exists}(v, f))$
  **else if** $(v \preceq v_r)$ **return** $R$
  **else return** $\mathsf{or}(t, f)$

Figure 2: Example ROBDD operations

$R_1 \lor R_2$ and $R_1 \leftrightarrow R_2$, $O(|R|)$ for $\neg R$, and $O(|R|^2)$ for $\exists v\ R$. Note however that we can test whether two ROBDDs are identical, whether $R_1 \leftrightarrow R_2$ is equivalent to true (1), in $O(1)$.

We give code for conjunction $R_1 \land R_2 = \mathsf{and}(R_1, R_2)$ and existential quantification (of one variable) $\exists v\ R = \mathsf{exists}(v, R)$. The code for disjunction $R_1 \lor R_2 = \mathsf{or}(R_1, R_2)$ is dual to $\mathsf{and}$ and is very similar in structure. The code make use of the auxiliary function $\mathsf{node}$ which builds a new ROBDD node. The $\mathsf{node}$ function returns $t$ if $t = f$, and is in practice is memoed so any call to $\mathsf{node}$ with the same arguments as a previous call returns a reference to the previously created ROBDD node.

Modern BDD packages provide many more operations, including specialized implementations of some operations for improved speed. Some important operations for our purposes are existential quantification of multiple variables $V$ for a formula $R\ \exists V\ R$, and the combination of conjunction and existential quantification $\exists V\ R_1 \land R_2$.

Although in theory the number of nodes in an ROBDDs can be exponential in the number of variables in the represented Boolean function, in practice ROBDDs are often very compact and computationally efficient. This is due to the fact that ROBDDs exploit to a very high-degree the symmetry of models of a Boolean formula.

## 3. Modeling Set CSPs Using ROBDDs

In this section we discuss how to solve set constraint satisfaction problems using ROBDDs. There are three parts to this—the modeling of set domains as ROBDDs, the modeling of set constraints as ROBDDs, and how to use both to produce a set solver.

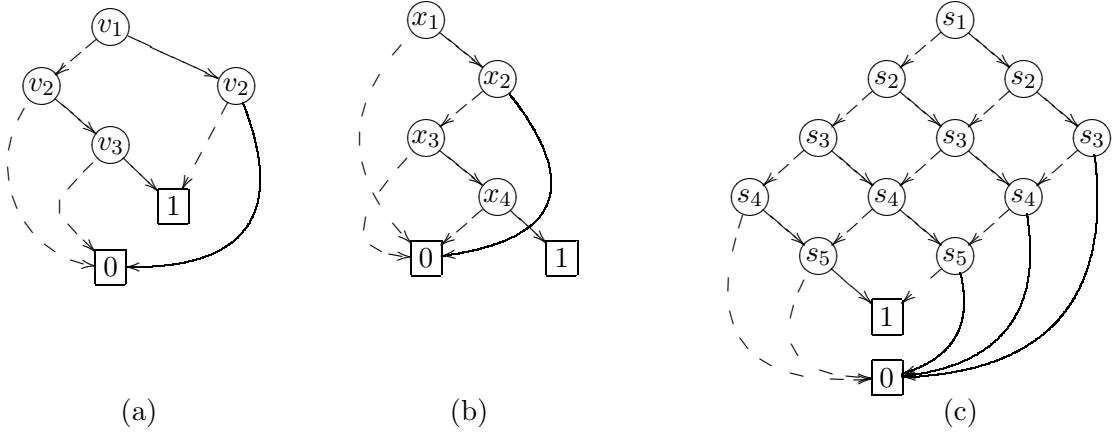

Figure 3: ROBDDs for (a) $\{\{1\}, \{1, 3\}, \{2, 3\}\}$, (b) $[\{1, 3, 4, 5\}, \{1, 3, 4, 5, 6, \ldots N\}]$ and (c) $\{s \subseteq \{1, 2, 3, 4, 5\} \mid |s| = 2\}$.

## 3.1 Modeling Set Domains using ROBDDs

Suppose the universe $\mathcal{U} = \{1, \ldots, N\}$. We assume that all set values are subsets of $\mathcal{U}$. Let $x$ be a set variable, and let $D$ be a domain over $\mathcal{V}$. Given that the size of the universe is bounded, we can associate a Boolean variable $x_i$ with each potential element $i \in \{1, \ldots, N\}$ of $x$. Hence a set variable $x$ is represented by a vector of Boolean variables $V(x) = \langle x_1, \ldots, x_N \rangle$.

Take any set $A \in D(x)$. Then we can represent $A$ as a valuation on the variables $\langle x_1, \ldots, x_N \rangle$ defined as $\theta_A = \{x_1 \mapsto (1 \in A), \ldots, x_n \mapsto (n \in A)\}$. We can represent this valuation $\theta_A$ and consequently the set $A$ as a Boolean formula $B(A)$ which has this valuation as its unique solution:

$$B(A) = \bigwedge_{i \in \mathcal{U}} y_i \text{ where } y_i = \begin{cases} x_i & \text{if } i \in A \\ \neg x_i & \text{otherwise} \end{cases}$$

Given that we can represent any element of $D(x)$ as a valuation, we can represent $D(x)$ itself as a Boolean formula $B(D(x))$ whose solutions $\theta_A$ correspond to the elements $A \in D(x)$. That is, $B(D(x))$ is the disjunction of $B(A)$ over all possible sets $A \in D(x)$:

$$B(D(x)) = \bigvee_{A \in D(x)} B(A) \text{ where } B(A) \text{ is defined above}$$

Any solution $\theta(x) \in D(x)$ then corresponds to a satisfying assignment of the Boolean formula $B(D(x))$. From now on we will overload the notion of a domain $D$ to equivalently return a set of possible sets $D(x)$ for a variable $x$, or its equivalent Boolean representation $B(D(x))$.

**Example 3.1.** Let $\mathcal{U} = \{1, 2, 3\}$. Suppose $v$ is a set variable with $D(v) = \{\{1\}, \{1, 3\}, \{2, 3\}\}$. We associate Boolean variables $\{v_1, v_2, v_3\}$ with $v$. We can then represent $D(v)$ as the Boolean formula $(v_1 \wedge \neg v_2 \wedge \neg v_3) \vee (v_1 \wedge \neg v_2 \wedge v_3) \vee (\neg v_1 \wedge v_2 \wedge v_3)$. The three solutions to this formula correspond to the elements of $D(v)$. The corresponding ROBDD $B(D(v))$ is shown in Figure 3(a).

This representation is useful since such Boolean formulæ can be directly represented as reasonably compact ROBDDs. Given a Boolean formula representing a domain, we can clearly construct the corresponding ROBDD in a bottom-up fashion, but in practice we only ever construct the ROBDD for a domain implicitly through constraint propagation. As we will show, ROBDDs permit a surprisingly compact representation of (many) subsets of $\mathcal{P}(\mathcal{U})$.

Since ROBDDs are ordered, we must specify a variable ordering for the Boolean variables that we use. We arbitrarily order the ROBDD variables for a single set variable $x$ as follows: $x_1 \prec x_2 \prec \ldots \prec x_n$. Assuming there is no special relationship between the elements of the universe, the choice of the ordering of ROBDD variables that represent a particular set variable is unimportant. However, the relative ordering of ROBDD variables comprising different set variables has a drastic effect on the size of the ROBDDs representing constraints, and is discussed in Section 3.2.

**Example 3.2.** An ROBDD representing the $2^{N-5}$ sets in the interval $[\{1,3,4,5\}, \{1,3,4,5,6,\ldots N\}]$ is shown in Figure 3(b).

The ROBDD representation is flexible enough to be able to represent compactly a wide variety of domains, even those that might at first seem difficult to represent. For example, the set of all subsets of $\{1,2,3,4,5\}$ of cardinality 2 can be represented by the ROBDD shown in Figure 3(c).

For convenience, we set all of the initial variable domains $D_{init}(x) = \mathcal{P}(\mathcal{U})$, which corresponds to the constant "1" ROBDD. Restrictions on the initial bounds of a domain can instead be expressed as unary constraints.

## 3.2 Modeling Primitive Set Constraints Using ROBDDs

A major benefit of using ROBDDs to model set constraint problems is that ROBDDs can be used to model the constraints themselves, not just the set domains. Any set constraint $c$ can be converted to a Boolean formula $B(c)$ on the Boolean variables comprising the set variables $vars(c)$, such that $B(c)$ is satisfied if and only if the corresponding set variable valuations satisfy $c$. As usual, we represent $B(c)$ as an ROBDD.

**Example 3.3.** Let $v$ and $w$ be set variables over the universe $\mathcal{U} = \{1,2,3\}$, and let $c$ denote the constraint $v \subseteq w$. Assume that the Boolean variables associated with $v$ and $w$ are $\langle v_1, v_2, v_3 \rangle$ and $\langle w_1, w_2, w_3 \rangle$ respectively. Then $c$ can be represented by the Boolean formula $(v_1 \rightarrow w_1) \wedge (v_2 \rightarrow w_2) \wedge (v_3 \rightarrow w_3)$. This formula can in turn be represented by either of the ROBDDs shown in Figure 4 (depending on the variable order).

Example 3.4 demonstrates that the order of the variables within the ROBDDs has a very great effect on the size of the formula representations.

**Example 3.4.** Consider again the constraint $v \subseteq w$ as described in Example 3.3. Figure 4 shows the effect of two different variable orderings on the size of the resulting BDD. In this

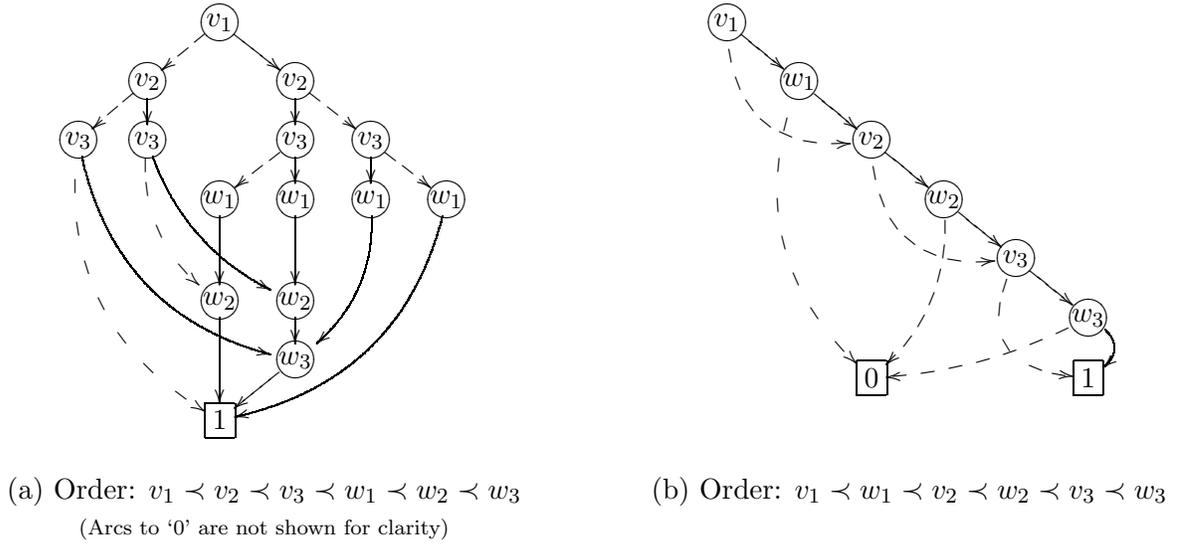

(a) Order: $v_1 \prec v_2 \prec v_3 \prec w_1 \prec w_2 \prec w_3$

(Arcs to '0' are not shown for clarity)

(b) Order: $v_1 \prec w_1 \prec v_2 \prec w_2 \prec v_3 \prec w_3$

Figure 4: Two ROBDDs for $v \subseteq w$

case, the variable ordering $v_1 \prec w_1 \prec v_2 \prec w_2$ gives a much more compact representation of constraint than the ordering $v_1 \prec v_2 \prec w_1 \prec w_2$. As can be seen from the figure, the latter ordering has size exponential in $n$, whereas the former has size linear in $n$.

If the variable set $\mathcal{V} = \{v_1, \ldots, v_m\}$ has corresponding Boolean variables $\langle v_{1,1}, v_{1,2}, \ldots, v_{1,N} \rangle$, $\langle v_{2,1}, v_{2,2}, \ldots, v_{2,N} \rangle$, $\ldots$, $\langle v_{m,1}, v_{m,2}, \ldots, v_{m,N} \rangle$, we choose to order the Boolean variables $v_{1,1} \prec \cdots \prec v_{m,1} \prec v_{1,2} \prec \cdots \prec v_{m,2} \prec \cdots \prec v_{1,N} \prec \cdots \prec v_{m,N}$. This ordering guarantees a linear representation for all of the primitive set constraints except cardinality. The reason is that primitive set constraints except cardinality are defined elementwise, that is an element $i \in v$ never constrains whether $j \in w$ or $j \notin w$ for $i \neq j$. For this reason there are no interactions between bits $v_i$ and $w_j$, $i \neq j$. Placing all the bits for the same element adjacent in the order means that the interactions of bits $v_i$ and $w_i$ are captured in a small ROBDD, which is effectively separate from the ROBDD describing the interactions of the next element's bits. Table 1 contains a list of the primitive set constraints, their corresponding Boolean formulæ and the sizes of the ROBDDs representing those formulæ under this point-wise ordering.

Cardinality constraints can be represented in a quadratic number of ROBDD nodes using a simple recursive definition. We define $card(\langle v_{i_1}, \ldots, v_{i_n} \rangle, l, u)$ to be the Boolean formula which restricts the number of true bits in the vector $\langle v_{i_1}, \ldots, v_{i_n} \rangle$ to between $l$ and

| $c$ | Boolean expression $B(c)$ | size of ROBDD |
|---|---|---|
| $k \in v$ | $v_k$ | $O(1)$ |
| $k \notin v$ | $\neg v_k$ | $O(1)$ |
| $u = d$ | $\bigwedge_{i \in d} u_i \wedge \bigwedge_{1 \le i \le N, i \notin d} \neg u_i$ | $O(N)$ |
| $u = v$ | $\bigwedge_{1 \le i \le N}(u_i \leftrightarrow v_i)$ | $O(N)$ |
| $u \subseteq v$ | $\bigwedge_{1 \le i \le N}(u_i \rightarrow v_i)$ | $O(N)$ |
| $u = v \cup w$ | $\bigwedge_{1 \le i \le N}(u_i \leftrightarrow (v_i \vee w_i))$ | $O(N)$ |
| $u = v \cap w$ | $\bigwedge_{1 \le i \le N}(u_i \leftrightarrow (v_i \wedge w_i))$ | $O(N)$ |
| $u = v - w$ | $\bigwedge_{1 \le i \le N}(u_i \leftrightarrow (v_i \wedge \neg w_i))$ | $O(N)$ |
| $u = \overline{v}$ | $\bigwedge_{1 \le i \le N} \neg(u_i \leftrightarrow v_i)$ | $O(N)$ |
| $u \ne v$ | $\bigvee_{1 \le i \le N} \neg(u_i \leftrightarrow v_i)$ | $O(N)$ |
| $|u| = k$ | $card(V(u), k, k))$ | $O(k(N-k))$ |
| $|u| \ge k$ | $card(V(u), k, N)$ | $O(k(N-k))$ |
| $|u| \le k$ | $card(V(u), 0, k)$ | $O(k(N-k))$ |

Table 1: Boolean representation of set constraints and the size of the corresponding ROBDD.

$u$ inclusive.

$$card(\langle v_{i_1}, \ldots, v_{i_n} \rangle, l, u) = \begin{cases} 1 & \text{if } l \le 0 \wedge n \le u \\ 0 & \text{if } n < l \vee u < 0 \\ (\neg v_{i_1} \wedge card(\langle v_{i_2}, \ldots, v_{i_n} \rangle, l, u)) \ \vee & \\ (v_{i_1} \wedge card(\langle v_{i_2}, \ldots, v_{i_n} \rangle, l-1, u-1) & \text{otherwise} \end{cases}$$

It is clear from the structure of $card(\langle v_{i_1}, \ldots, v_{i_n} \rangle, l, u)$ that the resulting ROBDD is $O(n^2)$ in size. A more general method of characterising cardinality constraints will be presented in Section 5.5.

### 3.3 A Basic Set Constraint Solver

We now show how to construct a simple set domain propagator $dom(c)$ for a constraint $c$. If $vars(c) = \{v_1, \ldots, v_n\}$, then we define a function $dom(c)$ mapping domains to domains as follows:

$$dom(c)(D)(v_i) = \begin{cases} \overline{\exists}_{V(v_i)} B(c) \wedge \bigwedge_{i=1}^{n} D(v_i) & \text{if } v_i \in vars(c) \\ D(v_i) & \text{otherwise} \end{cases}$$

In other words, to perform propagation we take the conjunction of the Boolean representations of the current domains of the variables in $vars(c)$ with the Boolean representation of the constraint $B(c)$ and project the result onto the Boolean variables $V(v_i)$ representing each variable $v_i$. Since $B(c)$ and all $D(v_i)$ are ROBDDs, this formula can be implemented directly using ROBDD operations.

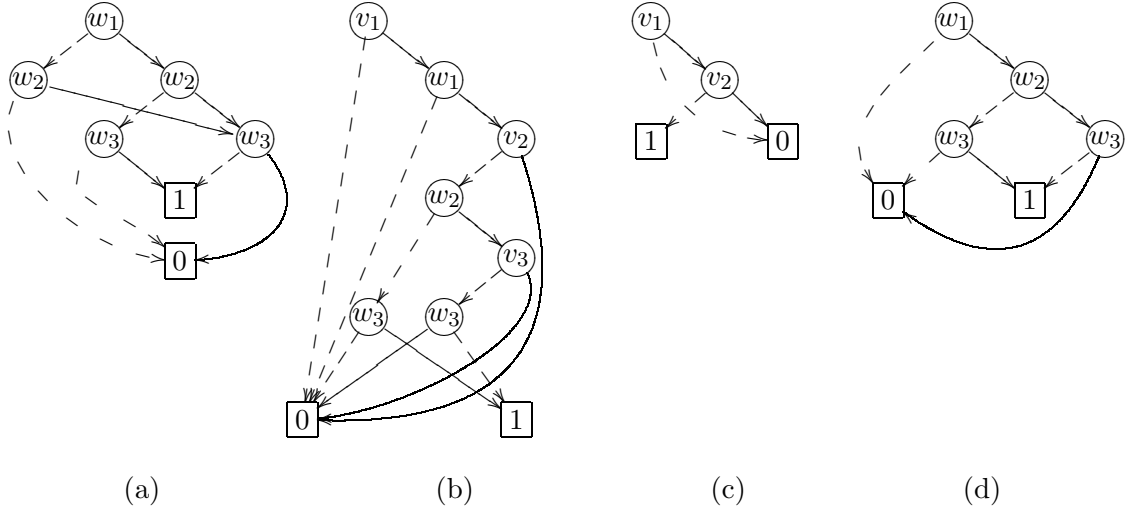

(a)          (b)          (c)          (d)

Figure 5: ROBDDs used in the domain propagation of $c \equiv v \subseteq w$ (a) $D(w)$, (b) $B(c) \wedge D(v) \wedge D(w)$, (c) $D'(v)$, and (d) $D'(w)$

---

**Example 3.5.** Consider the domain propagation of constraint $v \subseteq w$ with initial domain $D(v) = \{\{1\}, \{1,3\}, \{2,3\}\}$, and $D(w) = \{\{2\}, \{1,2\}, \{1,3\}\}$ from Examples 2.2 and 2.3. We conjoin the ROBDD $B(c)$ shown in Figure 4(b) with the domain ROBDD $D(v)$ shown in Figure 3(a) and the ROBDD for $D(w)$ shown in Figure 5(a). The result is an ROBDD representing solutions of the formula $v \subseteq w \wedge v \in D(v) \wedge w \in D(w)$ shown in Figure 5(b). We project the resulting onto $V(v)$ and $V(w)$, individually obtaining the ROBDDs $D'(v) = \{\{1\}, \{1,3\}\}$ and $D'(w) = \{\{1,2\}, \{1,3\}\}$ shown in Figure 5(c) and (d) respectively.

We need to verify the correctness of the propagator:

**Lemma 3.1.** *Let $\mathcal{V}$ be the collection of all set variables, and let $c$ be a set constraint with $vars(c) = \{v_1, \ldots, v_k\} \subseteq \mathcal{V}$. Then $dom(c)$ is a domain propagator for $c$.*

*Proof.* We need to verify that $dom(c)$ satisfies the identity of Definition 1. Suppose $D$ is a domain over $\mathcal{V}$. We need to check for each $v \in vars(c)$ that $\{char(\theta(v)) \mid \theta \in D \wedge \theta \in solns(c)\} = \{u \mid u \vDash \overline{\exists}_{V(v)} B(c) \wedge \bigwedge_{i=1}^{n} D(v_i)\}$, where $char(X)$ denotes the characteristic vector of $X$. This is clearly true, since the values $\theta \in solns(c)$ are by definition the satisfying assignments of $B(c)$, and the values $\theta \in D$ are by definition the satisfying assignments of $\wedge_{i=1}^{n} D(v_i)$. Hence the equality holds, implying that $dom(c)$ is a domain propagator. □

If $F$ is the set of all such domain propagators, then we can define a complete propagation algorithm $solv(F, D)$: For simplicity we use the set of constraints $C$ rather than the corresponding propagators.

  $\mathsf{solv}(C, D)$

$$Q := C$$
**while** $(\exists\, c \in Q)$
$\qquad D' := dom(c)(D)$
$\qquad$ **if** $(|\,vars(c)| = 1)\ C := C - \{c\}$
$\qquad V := \{v \in \mathcal{V} \mid D(v) \neq D'(v)\}$
$\qquad Q := (Q \cup \{c' \in C \mid\ vars(c') \cap V \neq \emptyset\}) - \{c\}$
$\qquad D := D'$
$\quad$ **return** $D$

We maintain a queue $Q$ of constraints to be (re-)propagated, initially all $C$. We select a constraint $c$ from the queue to propagate, calculating a new domain $D'$. If the constraint is unary we remove it from $C$, never to be considered again, since all information is captured in the domain. We then determine which variables $V$ have changed domain, and add to the queue constraints $c' \in C$ involving these variables, with the exception of the current constraint $c$.

By combining this algorithm with the modeling techniques described earlier, we have shown how to construct a simple ROBDD-based set domain propagator. Various improvements to this basic scheme will be discussed in Section 4 and Section 6.

The reader may wonder whether in this section we have done anything more than map set constraints to Boolean constraints and then apply a Boolean ROBDD solver to them. This is *not* the case. Crucially the domain propagation is on the original set variables, not on the Boolean variables that make them up. In fact the set bounds consistency approach (Puget, 1992; Gervet, 1997) can be considered as simply a mapping of set constraints to Boolean constraints.

## 4. Effective Modeling of Set Constraints Using ROBDDs

In this section we demonstrate that the ROBDD-based modeling approach is very flexible, allowing us to produce highly efficient implementations of many complex constraints.

### 4.1 Combining Constraints and Removing Intermediate Variables

It is rarely possible to express all of the constraints in a real set problem directly as primitive set constraints on the variables of the original problem. Instead, we would usually like to express more complicated set constraints, which can be decomposed into multiple primitive set constraints, often requiring the introduction of intermediate variables.

**Example 4.1.** Let $c$ be the constraint $|v \cap w| \leq k$, which requires that $v$ and $w$ have at most $k$ elements in common. This constraint is used in modeling the problem of finding Steiner systems (see Section 7.1). Since this is not a primitive set constraint, existing set solvers would usually implicitly decompose it into two primitive set constraints and introduce an intermediate variable $u$. The representation of the constraint then becomes $\exists u\ u = v \cap w \wedge |u| \leq k$.

In the case of Example 4.1 the decomposition does not affect the strength of the resulting propagator. To prove this fact, we use two results presented by Choi, Lee, and Stuckey (2003). Choi, Lee, and Stuckey prove these results in the case of a finite integer domain solver, although the proofs in the set domain case are identical. [1]

**Lemma 4.1 (Choi et al., 2003).** *Let $c_1$ and $c_2$ be set constraints. Then $solv(\{dom(c_1 \wedge c_2)\}, D) \sqsubseteq solv(\{dom(c_1), dom(c_2)\}, D)$ for all domains $D$.*

**Lemma 4.2 (Choi et al., 2003).** *Let $c_1$ and $c_2$ be two set constraints sharing at most one variable $x \in \mathcal{V}$. Then $solv(\{dom(c_1), dom(c_2)\}, D) = solv(\{dom(c_1 \wedge c_2)\}, D)$ for all domains $D$.*

Even if the strength of the propagator is unaffected by the decomposition, splitting a propagator introduces a new variable, thus slowing down the propagation process. In cases where two set constraints share more than one variable Lemma 4.2 does not apply, so in such a case there can also be a loss of propagation strength.

The ROBDD representation of the constraints allows us to utilise such complex constraints directly, thus avoiding the problems associated with splitting the constraint. We can directly construct an ROBDD for complex constraints by forming the conjunction of the corresponding primitive constraints and existentially quantifying away the intermediate variables.

**Example 4.2.** Consider the constraint $c \equiv |v \cap w| \leq k$ as discussed in Example 4.1. We can build a domain propagator $dom(c)$ directly for $c$ by constructing the ROBDD $\exists V(u)\ u = v \cap w \wedge |u| \leq k$. In this case the size of the resulting ROBDD is $O(kN)$.

The ROBDD for $|v \cap w| \leq 2$, that is $\exists V(u)\ B(u = v \cap w) \wedge B(|u| \leq 2)$, where $\mathcal{U} = \{1, 2, 3, 4, 5\}$ is shown in Figure 6(a). The ROBDD for $|u| \leq 2$ is shown in Figure 6(b) for comparison. Note how each $u_i$ node in the Figure 6(b) is replaced by $v_i \wedge w_i$ (the formula for $u_i$) in Figure 6(a).

## 4.2 Modeling Global Constraints

As the previous section demonstrates, it is possible to join the ROBDDs representing primitive set constraints into a single ROBDD representing the conjunction of the constraints. We can use this to join large numbers of primitive constraints to form *global constraints*, which in many cases will improve performance due to stronger propagation.

It is one of the strengths of the ROBDD-based modeling approach that it is trivial to construct global constraints simply using ROBDD operations on primitive set constraints, without laboriously writing code to perform propagation on the global constraint. This approach is very powerful, but is not feasible for all combinations of primitive constraints. As we shall see, some global constraints that we might desire to construct lead to ROBDDs that are exponentially large.

A useful global constraint is the constraint `partition`$(v_1, \ldots, v_n)$, which requires that the sets $v_1, v_2, \ldots, v_n$ form a partition of the universe $\mathcal{U} = \{1, \ldots, N\}$. We can easily

---

[1]. The observation of Choi et al. that Lemma 4.2 does not apply if the shared variable is a set variable is only true if we are performing set bounds propagation, not set domain propagation.

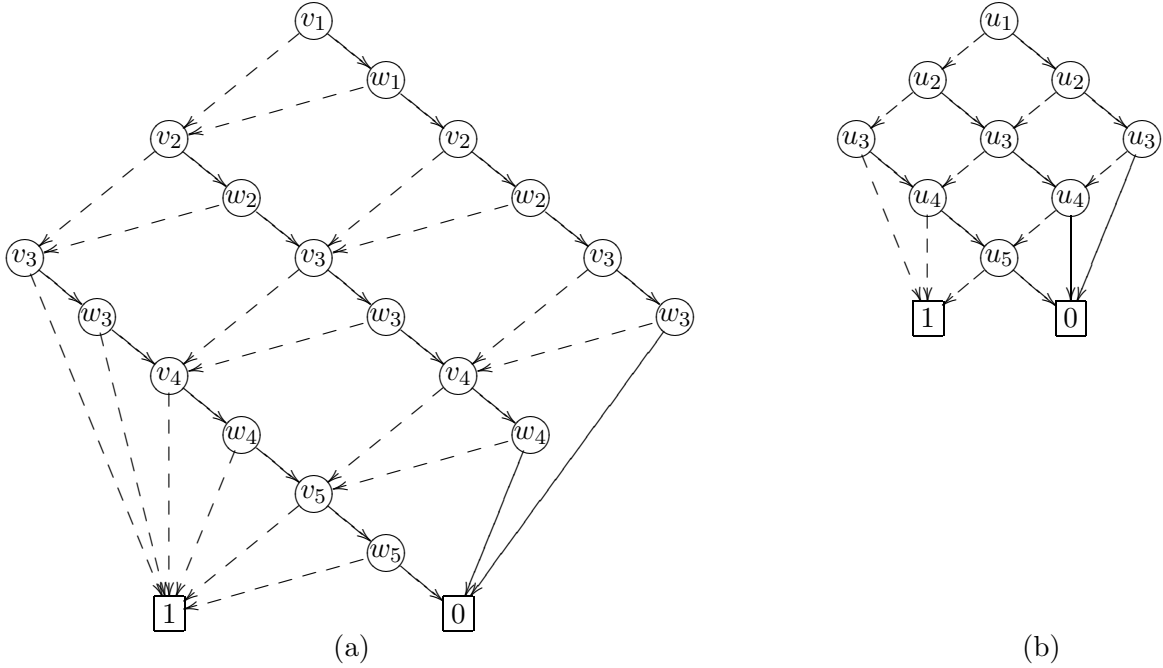

<div style="text-align: center">(a)</div>
<div style="text-align: right">(b)</div>

Figure 6: ROBDDs for (a) $|v \cap w| \leq 2$ where $v, w \subseteq \{1, 2, 3, 4, 5\}$, and (b) $|s| \leq 2$ where $s \subseteq \{1, 2, 3, 4, 5\}$.

construct this constraint from primitive set constraints as follows:

$$\texttt{partition}(v_1, \ldots, v_n) = \bigwedge_{i=1}^{n-1} \bigwedge_{j=i+1}^{n} \exists_{u_{ij}} (u_{ij} = v_i \cap v_j \wedge u_{ij} = \phi)$$

$$\wedge \quad \exists_{w_0} \cdots \exists_{w_n} (w_0 = \phi \wedge (\bigwedge_{i=1}^{n} w_i = w_{i-1} \cup v_i) \wedge w_n = \mathcal{U})$$

The propagator $dom(\texttt{partition}(\cdots))$ is stronger than a domain propagator on the decomposition. For example, consider the constraint $c \equiv \texttt{partition}(x, y, z)$ as depicted in Figure 7, and the domain $D$ where $D(x) = D(y) = \{\{1\}, \{2\}\}$ and $D(z) = \{\{2\}, \{3\}\}$. Domain propagation using the decomposition of the constraint will not alter $D$, but domain propagation using the global constraint will give $dom(c)(D)(z) = \{\{3\}\}$. Hence stronger propagation can be gained from global propagation using this constraint.

Unfortunately not all global constraints can be modelled efficiently using this approach. In particular, there is a risk that the ROBDD representation of a global constraint could be extremely large, making it infeasible to construct and use for propagation.

For example, consider the constraint $\texttt{atmost}(\langle v_1, \ldots, v_n \rangle, k)$ proposed by Sadler and Gervet (2001), which requires that each of the sets $v_1, \ldots, v_n$ has cardinality $k$ and the intersection of any pair of the sets is at most 1 element in size. This constraint models the

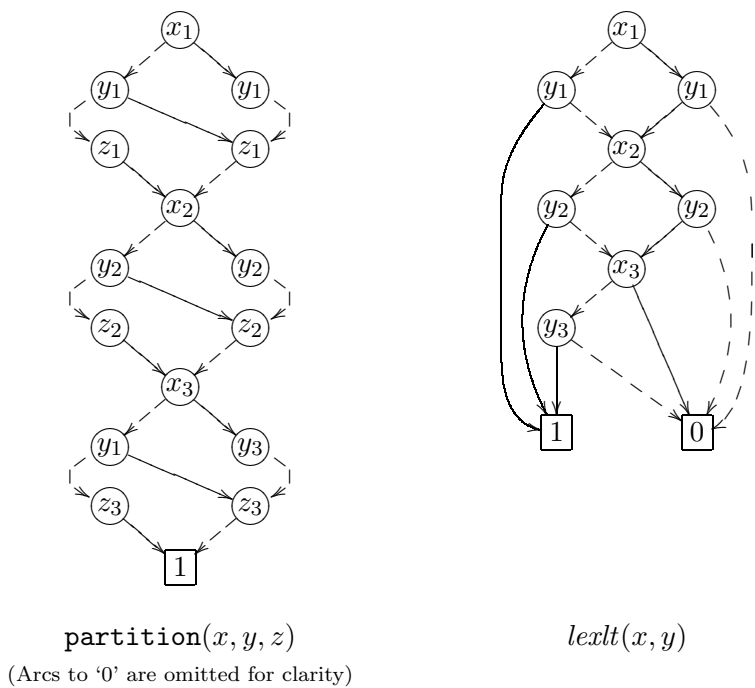

partition$(x, y, z)$            *lexlt*$(x, y)$

(Arcs to '0' are omitted for clarity)

Figure 7: ROBDDs for the constraints partition$(x, y, z)$ and *lexlt*$(x, y)$, where $x$, $y$ and $z$ are set variables taking values from the universe $\mathcal{U} = \{1, 2, 3\}$.

Steiner triple systems of Section 7.1. Bessiere, Hebrard, Hnich, and Walsh (2004) proved that enforcing bounds consistency on this constraint is NP-hard, so it follows that enforcing domain consistency on this constraint is at least NP-hard as well. Theoretically we can still construct an ROBDD representing this constraint from primitive constraints as follows:

$$\bigwedge_{i=1}^{n} |v_i| = n \;\wedge\; \bigwedge_{i=1}^{n-1} \bigwedge_{j=i+1}^{n} \exists_{u_{ij}} (u_{ij} = v_i \cap v_j \wedge |u_{ij}| \leq 1)$$

Unfortunately, the resulting ROBDD turns out to be exponential in size, making it impractical for use as a global propagator. This is not surprising in light of NP-hardness of the problem in general—in fact it would be surprising if the resulting ROBDD was *not* exponential in size!

## 4.3 Avoiding Symmetry Through Ordering Constraints

It is important in modeling a constraint satisfaction problem to minimize symmetries in the model of the problem. A model that contains symmetrical solutions often has a greatly enlarged search space, leading to large amounts of time being spent in searching sections of a search tree that are identical up to symmetric rearrangement. It is therefore highly desirable to remove whatever symmetries exist in a problem.

One approach to symmetry-breaking is the introduction of additional ordering constraints between the variables of the problem. A convenient ordering to use on sets is a *lexicographic order* on the characteristic bit vectors of the sets. In other words, if $v$ and $w$ are set variables, then $v < w$ in the lexicographic ordering if and only if the list of bits $V(v)$ is lexicographically smaller than $V(w)$. We can model this lexicographic ordering constraint as $lexlt(v, w, 1)$, which is defined recursively in the following manner:

$$lexlt(v, w, n) = \begin{cases} 0 & \text{if } n > N \\ (\neg v_n \wedge w_n) \vee ((v_n \leftrightarrow w_n) \wedge lexlt(v, w, n+1)) & \text{otherwise} \end{cases}$$

An example of such an ROBDD is depicted in Figure 7.

We shall make use of the lexicographic ordering constraint extensively in our experiments in Section 7.

## 5. Modeling Integers, Multisets, and Weighted Sum Constraints

In this section we show how to model integer variables and integer constraints using ROBDDs. In general such representations can be very large, which limits the usefulness of ROBDDs as the basis for a general purpose finite-domain constraint solver. Despite this, the ability to represent integers is extremely useful as a component of a set solver. We propose two major uses of the integer representation.

Firstly, we can use the integer representation to model values such as the weighted sum of the elements of a set. Constraints on the weighted sum of elements of a set have been shown to be useful in practical applications (Mailharro, 1998).

Secondly, we can model finite multisets using ROBDDs by replacing the individual ROBDD variables of the set representation with bundles of ROBDD variables, each bundle corresponding to a binary integer. Multiset operations can then be constructed by composing integer operations on the variable bundles.

In addition, the integer representation as described here could be used as an interface between an ROBDD-based set solver and a more conventional integer finite-domain solver. Such an interface could easily be implemented using channeling constraints between ROBDD integer and finite-domain versions of the same variable.

### 5.1 Representing Integer Values using ROBDDs

In order to model integers and integer operations we must choose an appropriate representation in terms of Boolean formulæ. In general we are free to use any encoding of an integer as a binary sequence, such as unary, unsigned binary and twos-complement encodings, but for simplicity we choose to represent all of our integers in unsigned binary form.

We can represent an arbitrary integer expression $e$ by a list of Boolean formulæ. Each formula in the list corresponds to a single bit of the unsigned binary value of the expression. We will denote such a list by $\langle e_{n-1}, e_{n-2}, \ldots, e_1, e_0 \rangle$, where each $e_i$ is a Boolean formula, in order from the most significant bit to the least significant bit. We interpret a formula as a "1" bit if the formula is logically true, and as a "0" bit otherwise; for simplicity we will just call the formulæ the "bits of the expression". We will also use the expression $e$ and its list of constituent bits interchangeably. As usual, we will represent the Boolean formulæ as ROBDDs. As we shall see, this notation is flexible enough to represent arbitrary integer expressions.

**Example 5.1.** Consider the integer constant $k = 25$, which is 11001 in unsigned binary. We will represent $k$ as the list $\langle 1, 1, 0, 0, 1 \rangle$.

In order to represent an integer variable $x$, we associate with $x$ a fixed set of Boolean variables $\{x_{k-1}, \ldots, x_0\}$. The value of $x$ is then taken to be the value of the unsigned binary integer $\langle x_{k-1}, \ldots, x_0 \rangle$. By varying the value of the $x_i$ variables, the value of $x$ can range from 0 to $2^k - 1$ inclusive. As always the ordering of the Boolean variables has a significant effect on the sizes of the ROBDD representations of formulæ. If $x = \langle x_{k-1}, \ldots, x_0 \rangle$, $y = \langle y_{k-1}, \ldots, y_0 \rangle$, and $z = \langle z_{k-1}, \ldots, z_0 \rangle$, then we choose to order the corresponding ROBDD variables in a interleaved most-significant-bit-first order, i.e. $x_{k-1} \prec y_{k-1} \prec z_{k-1} \prec x_{k-2} \prec \cdots \prec x_0 \prec y_0 \prec z_0$.

Since we permit arbitrary Boolean formulæ as the bits of an expression, we can also model arbitrary integer expressions. For example, suppose $x$ and $y$ are integer variables with bits $\langle x_2, x_1, x_0 \rangle$ and $\langle y_2, y_1, y_0 \rangle$ respectively. Then, for example, we can represent the expression $x \wedge y$ (where the $\wedge$ denotes the bitwise AND operator) as the list $\langle x_2 \wedge y_2, x_1 \wedge y_1, x_0 \wedge y_0 \rangle$. We are not limited to logical operations, as we shall see in the next section.

## 5.2 Representing Integer Operations using ROBDDs

We can also use ROBDDs to model arithmetic operations such as addition, by analogy with the design of the corresponding logic circuits. For the purpose of the set and multiset solvers, we only require implementations of the operations of addition, left shift, minimum, maximum, multiplication by a constant, and multiplication by a single variable bit. We do not require a general implementation of multiplication using ROBDDs.

It is convenient to assume that all integer expressions have the same number of bits. We may assume this without loss of generality since we can freely pad the left of the shorter of any pair of expressions with "0" bits.

To model addition, we simulate the operation of a full binary adder. Suppose $x$ and $y$ are integer expressions, with bit representations $\langle x_{l-1}, \ldots, x_0 \rangle$ and $\langle y_{l-1}, \ldots, y_0 \rangle$. We can use ROBDDs to compute the output bits $plus(x, y)$ of the operation $x + y$ as follows (here $c_i$ denotes a carry bit and $s_i$ denotes a sum bit):

$$
\begin{aligned}
c_{-1} &= 0 \\
s_i &= x_i \oplus y_i \oplus c_{i-1} & 0 \leq i < l \\
c_i &= (\neg c_{i-1} \wedge x_i \wedge y_i) \vee (c_{i-1} \wedge (x_i \vee y_i)) & 0 \leq i < l \\
plus(x, y) &= \langle c_{l-1}, s_{l-1}, \ldots, s_1, s_0 \rangle
\end{aligned}
$$

Note we avoid overflow by extending the size of the result by one bit.

**Example 5.2.** Suppose $x$ is an integer variable, with bits $\langle x_1, x_0 \rangle$. We can represent the expression $x + 3$ by the bits $\langle (\neg x_0 \wedge x_1) \vee x_0), x_1 \oplus 1 \oplus x_0, x_0 \oplus 1 \rangle = \langle x_0 \vee x_1, x_0 \leftrightarrow x_1, \neg x_0 \rangle$.

The left shift operation is trivial to implement. If $x$ is an integer expression represented by $\langle x_{l-1}, \ldots, x_0 \rangle$ and $k$ is a non-negative integer, then we can represent the left shift of $x$

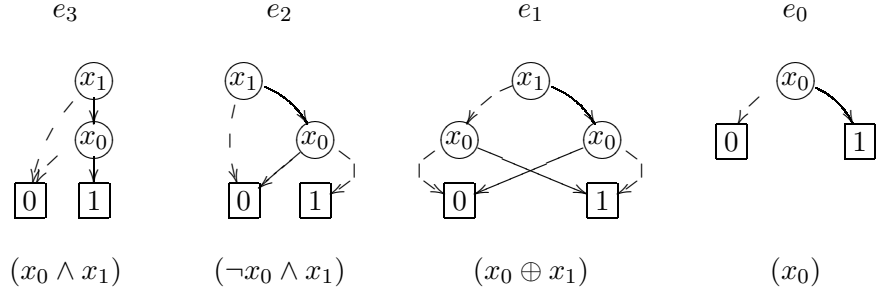

Figure 8: ROBDDs representing the bits $\langle e_3, e_2, e_1, e_0 \rangle$ of the expression $e = mul(\langle x_1, x_0 \rangle, 3)$, together with the simplified Boolean expressions for each $e_i$.

---

by $k$ bits $shl(x, k)$ by the following formula:

$$shl(\langle x_{l-1}, \ldots, x_0 \rangle, k) = \langle x_{l-1}, \ldots, x_0, \underbrace{0, \ldots, 0}_{\text{k bits}} \rangle$$

We can implement the operation of multiplication by a constant using the *plus* and *shl* operators. If $x$ is an integer expression with bits $\langle x_{l-1}, \ldots, x_0 \rangle$ and $k$ is a non-negative integer, then $x \times k$ corresponds to $mul(x, k)$ in the following formula:

$$mul(x, k) = \begin{cases} \langle \rangle & \text{if } k = 0 \\ mul(shl(x, 1), \frac{k}{2}) & \text{if } k \text{ is even and } k > 0 \\ plus(x, mul(shl(x, 1), \lfloor \frac{k}{2} \rfloor)) & \text{if } k \text{ is odd and } k > 0 \end{cases} \quad (1)$$

**Example 5.3.** Let $x = \langle x_1, x_0 \rangle$, and consider the expression $e = mul(x, 3)$. By applying Equation (1), we obtain $e = \langle x_1 \wedge (x_1 \wedge x_0), x_1 \oplus (x_0 \wedge x_1), x_0 \oplus x_1, x_0 \rangle$ which can be simplified to $\langle x_0 \wedge x_1, \neg x_0 \wedge x_1, x_0 \oplus x_1, x_0 \rangle$. The corresponding ROBDD representations are shown in Figure 8.

## 5.3 Integer Constraints using ROBDDs

We can also express constraints on integer expressions using ROBDDs. In particular, we show how to implement equality and inequality constraints. As usual, we assume any two expressions $x$ and $y$ have equal lengths of $l$ bits; if not, we pad the shorter expression with "0" bits on the right.

Equality of two integer expressions $x$ and $y$ is easy to represent as an ROBDD—the corresponding bits of $x$ and $y$ must be equal. Hence we can represent the equality constraint $x = y$ as the ROBDD $B(x = y)$:

$$B(x = y) = \bigwedge_{i=0}^{l-1} x_i \leftrightarrow y_i$$

Note that this is identical to the implementation of equality for two set expressions with the addition of zero-padding.

It turns out that we already have an implementation for inequality constraints, albeit in a different guise. Inequalities of the binary integers correspond to inequalities on the lexicographic ordering of the bit representations, so we can implement, for example, the strict less-than constraint $x < y$ for two integer variables $x$ and $y$ using the *lexlt* operation from Section 4.3.

$$B(x < y) = lexlt(x, y)$$

We can then use this to construct the reverse inequality by swapping the order of the operands, and the non-strict inequality by negating the formula (reversed as necessary).

The implementation of inequalities also leads us to an implementation of the minimum and maximum expressions. Consider the problem of finding the smaller of two integer expressions $x$ and $y$. If $x$ and $y$ have bit vectors $\langle x_{l-1}, \ldots, x_0 \rangle$ and $\langle y_{l-1}, \ldots, y_0 \rangle$ respectively, we recursively define $\min(x, y)$ as follows:

$$
\begin{aligned}
L_l &= R_l = 0 \\
L_i &= L_{i+1} \vee (\neg L_{i+1} \wedge \neg R_{i+1} \wedge \neg x_i \wedge y_i) & 1 \leq i < l \\
R_i &= R_{i+1} \vee (\neg L_{i+1} \wedge \neg R_{i+1} \wedge x_i \wedge \neg y_i) & 1 \leq i < l \\
m_i &= (L_{i+1} \wedge x_i) \vee (R_{i+1} \wedge y_i) \vee (\neg L_{i+1} \wedge \neg R_{i+1} \wedge x_i \wedge y_i) & 0 \leq i < l \\
\min(x, y) &= \langle m_{l-1}, \ldots, m_0 \rangle
\end{aligned}
$$

In the equation above, the $L_i$ and $R_i$ values are flag bits which state whether the higher order bits have already allowed us to conclude that one of the two values is the minimum. The maximum operation is defined similarly.

### 5.4 Modeling Multisets and Multiset Constraints

Various authors have suggested that multisets are a valuable addition to the modeling abilities of a set constraint solver (Kiziltan & Walsh, 2002). In this section we briefly show how multisets and multiset constraints can be modelled using ROBDDs by making use of the integer building blocks described above.

A multiset $m$ is a unordered list of elements $\{\!\{ m_0, \ldots, m_n \}\!\}$ drawn from the universe $\mathcal{U}$, in which (unlike a set) repetition of elements is permitted. Most set operations have parallel operations on multisets, although the multiset operations are not strict generalisations of the set operations. Let $occ(i, m)$ denote the number of occurrences of an element $i$ in a multiset $m$. Suppose $m$ and $n$ are multisets, and $k$ is an integer constant. We define the following multiset relations and operations by their actions on the number of occurrences of each element $i$ in the universe:

- Equality: $m = n$ iff $occ(i, m) = occ(i, n)$ for all $i \in \mathcal{U}$.

- Subset: $m \subseteq n$ iff $occ(i, m) \leq occ(i, n)$ for all $i \in \mathcal{U}$.

- Union: $occ(i, m \cup n) = occ(i, m) + occ(i, n)$ for all $i \in \mathcal{U}$.

- Intersection: $occ(i, m \cap n) = \min\{occ(i, m), occ(i, n)\}$ for all $i \in \mathcal{U}$.

- Difference: $occ(i, m \setminus n) = \max\{0, occ(i, m) - occ(i, n)\}$ for all $i \in \mathcal{U}$.

- Cardinality: $|m| = \sum_{i \in \mathcal{U}} occ(i, m)$ for all $i \in \mathcal{U}$.

To represent a set variable $x$ we associated a vector of Boolean variables $\langle x_1, \ldots, x_n \rangle$ with the bits of the characteristic vector of a valuation of $x$. In the case of a multiset, the characteristic vector of a multiset $m$ is a vector of integers, and so we need to associate an integer value $m_i$ with each potential element $i$ of the multiset $m$. We can model each such integer value using the approach described above.

If $m$ is a multiset variable, then we associate a *bundle* of ROBDD variables with every $i \in \mathcal{U}$, the contents of which comprise the bits of the corresponding integer expression $m_i$. In order that we can represent multisets in a finite number of bits, we assume that the number of occurrences of any element in a multiset variable is bounded above by a reasonably small $M$, allowing us to use only $k = \lceil \log_2 M \rceil$ Boolean variables per bundle. We then write $m$ as a list of bundles $\langle m_1, \ldots, m_n \rangle$, where each $m_i$ is in turn a list of bits $\langle m_{i,k-1}, \ldots, m_{i,0} \rangle$.

Given a representation of multiset variables, we now turn our attention to the implementation of multiset expressions and constraints. Multiset expressions can be implemented in the obvious way as sequences of integer expressions. For example, suppose $x$ and $y$ are multiset variables with associated bundles $\langle x_1, \ldots, x_N \rangle$ and $\langle y_1, \ldots, y_N \rangle$ respectively. Then the bundles corresponding to the expression $x \cup y$ are $\langle plus(x_1, y_1), plus(x_2, y_2), \ldots, plus(x_N, y_N) \rangle$. Similarly, the bundles corresponding to the expression $x \cap y$ are $\langle \min(x_1, y_1), \ldots, \min(x_N, y_N) \rangle$, and so on for other expressions. We show how to implement cardinality and weighted sum constraints in Section 5.5.

Multiset constraints are also trivial to implement—for example, two multisets $x$ and $y$ are equal if and only if all of their constituent bundles are equal, and so multiset equality can be modelled by a conjunction of integer equalities. Relations such as subset correspond to a conjunction of integer inequalities on the constituent bundles, the implementation of which was described in Section 5.3.

Up until this point we have left the ordering of the ROBDD variables that make up a multiset variable unspecified. Unfortunately, unlike the set case, there is no single optimal variable ordering that is guaranteed to produce compact descriptions of all the primitive multiset constraints. For example, a subset constraint can be compactly represented by a bundle-major bit ordering but not a bit-major ordering (see Figure 9), since a subset constraint consists of a series of integer inequalities between the corresponding bundles of two multiset variables, and so a bundle-major ordering gives an interleaving of variables as above. However, the opposite is true for a cardinality constraint, which consists of a sum of the values of the bundles within a variable, the variables of which are interleaved under a bit-major ordering but not a bundle-major ordering. These two orderings are mutually exclusive, and hence we can conclude that in general there need not be an optimal variable ordering for modeling multiset constraints.

## 5.5 Weighted Sum and Cardinality Constraints

In many practical applications we are interested in placing constraints on a weighted sum of the elements of a set variable. For example, in the Balanced Academic Curriculum problem (problem `prob030` of CSPLib), every course has an associated weight corresponding to its academic load, and there is a limit to the total academic load that can be undertaken during any given time period. In the set model of the problem proposed by Hnich, Kiziltan, and Walsh (2002), the limits on the academic load in a period are made through the use of a weighted sum constraint. In addition the cardinality constraint which we have already described is a special case of the weighted sum constraint where all the weights are set to "1". It therefore seems essential to implement this constraint in the ROBDD framework.

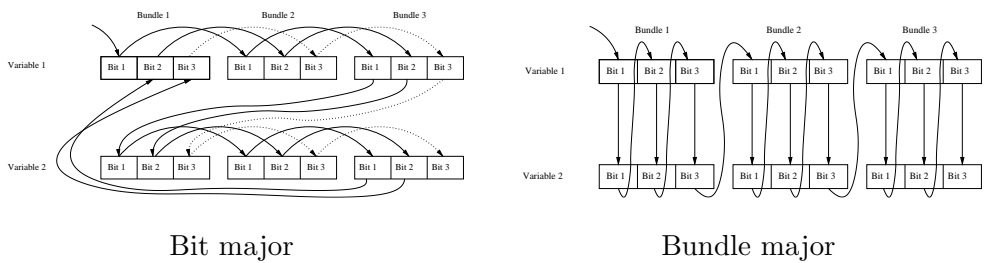

Figure 9: Bit and bundle major orderings for two multiset variables

Suppose $x$ is a set or multiset expression with bit bundles $\langle x_1, \ldots, x_n \rangle$. (If $x$ is a set expression, then each bit bundle has size 1). We can use the integer operations described earlier to produce an integer expression $\mathtt{wsum}(x, w)$ corresponding to the weighted sum $\sum_{i=1}^{n} x_i w_i$, where $w$ is a vector of integers $\langle w_1, \ldots, w_n \rangle$:

$$\phi_0 = \langle \rangle$$
$$\phi_i = plus(\phi_{i-1}, mul(x_i, w_i))$$
$$\mathtt{wsum}(x, w) = \phi_n$$

Expressions involving the cardinality of set and multiset variables can then be expressed as a special case of the weighted sum expression where $w_i = 1$ for all $1 \leq i \leq n$. We have already seen one method of constructing ROBDDs for the constraints $|x| = k$, $|x| \leq k$ and $|x| \geq k$ in the set variable case; this method is more general, since it permits us to directly model constraints such as $|x| + 5 \leq |y| + |z|$. This is of great practical value—for example it is needed to implement the Hamming Code experiment of Section 7.3. In the cases already discussed in Section 3.2 the ROBDDs produced by the two methods are identical since ROBDDs are a canonical representation.

## 6. Efficient Constraint Propagation Using ROBDDs

In this section we discuss improvements to and variants on the basic domain propagation scheme presented in Section 3.3.

The implementation of the domain propagator $dom(c)$ has a substantial effect on the performance of the solver. The definition of $dom(c)$ was given in its purest mathematical form for simplicity. Section 6.1 discusses several implementation details that can lead to greatly improved efficiency when performing domain propagation.

In general, the inferences that can be obtained using the domain propagator may be too costly to be practical, and so in some circumstances it may be desirable to enforce a weaker form of consistency. Weaker consistency may be substantially cheaper to enforce, permitting more time to be spent in searching the solution space.

As always there is a compromise between propagation time and search time, and for certain problems it may be more productive to spend more time searching than performing more accurate propagation, while for other problems the converse may be true. The ROBDD based representation allows this to be taken to extremes—in theory it is possible to form a single ROBDD representing the solutions to a constraint satisfaction problem by forming the ROBDD conjunction of the constraints. A solution could then be trivially read from the ROBDD as a satisfying assignment. Usually such an ROBDD would be prohibitively expensive to construct in both time and space, forcing us to maintain less strict consistency levels.

Accordingly, we show how to implement weaker levels of consistency using the ROBDD representation by combining the domain propagator with an approximation operation. The approximation operation simplifies the ROBDD representing a domain to its bounds closure under a suitable definition of bounds. The ROBDD representing the bounds of a domain is almost always smaller than the domain itself, leading to better performance for all future operations on that domain. As we shall see, this can lead to substantial performance improvements for the overall solver.

## 6.1 Domain Propagation

Let $c$ be a constraint, with $vars(c) = \{v_1, \ldots, v_n\}$. Section 3.3 gave the following definition of a domain propagator:

$$dom(c)(D)(v_i) = \overline{\exists}_{V(v_i)}(B(c) \wedge \bigwedge_{j=1}^{n} D(v_j)) \qquad (2)$$

Since $B(c)$ and $D(v_j)$ are ROBDDs, we can directly implement Equation (2) using ROBDD operations. In practice it is more efficient to perform the existential quantification as early as possible to limit the size of the intermediate ROBDDs. Here we make use of the efficient combined conjunction and existential quantification operation, which we'll call *and-abstraction*, provided by most ROBDD packages.

This leads to the following implementation:

$$
\begin{aligned}
\phi_i^0 &= B(c) \\
\phi_i^j &= \begin{cases} \exists_{V(v_j)}(D(v_j) \wedge \phi_i^{j-1}) & 1 \le i, j \le n, \ i \ne j \ \text{(and-abstraction)} \\ \phi_i^{i-1} & i = j \end{cases} \\
dom(c)(D)(v_i) &= D(v_i) \wedge \phi_i^n
\end{aligned}
\qquad (3)
$$

The worst case complexity is still $O(|B(c)| \times \prod_{j=1}^{n} |D(v_j)|)$ for each $v_j$. Clearly some of the computation can be shared between propagation of $c$ for different variables since $\phi_i^j = \phi_{i'}^j$ when $j < i$ and $j < i'$. Even with this improvement the algorithm of Equation (3) uses $O(n^2)$ and-abstraction operations (which experimentally have been shown to occupy the majority of the execution time of the set solver).

The domain propagator implementation of Equation (3) can be significantly improved by observing that in the case of an $n$-variable constraint ($n \ge 3$) many similar sub-formulæ

dom_divide_conquer($D$, $\phi$, $V$):
    **if** $(|V| = 0)$ **return** $D$
    **else if** $(V = \{v_i\})$
        $D(v_i) := D(v_i) \wedge \phi$
        **return** $D$
    **else**
        $\{v_1, \ldots, v_k\} := V$
        $h := \lfloor \frac{k}{2} \rfloor$
        $R := \exists_{V(v_1)} D(v_1) \wedge \exists_{V(v_2)} D(v_2) \wedge \cdots \exists_{V(v_h)} D(v_h) \wedge \phi$
        $L := \exists_{V(v_{h+1})} D(v_{h+1}) \wedge \exists_{V(v_{h+2})} D(v_{h+2}) \wedge \cdots \exists_{V(v_k)} D(v_k) \wedge \phi$
        $D_1 :=$ dom_divide_conquer($D$, $L$, $\{v_1, \ldots, v_h\}$)
        $D_2 :=$ dom_divide_conquer($D_1$, $R$, $\{v_{h+1}, \ldots, v_k\}$)
        **return** $D_2$

Figure 10: A divide and conquer algorithm for domain propagation

are computed. Due to the need to perform the existential quantification operations as early as possible, we do not have complete freedom to rearrange the order of evaluation as we see fit. However, a simple divide-and-conquer strategy for calculating $dom(c)(D)(v_i)$ allows us to perform domain propagation using just $O(n \log n)$ and-abstraction operations. We define $dom(c)(D) =$ dom_divide_conquer($D$, $B(c)$, $vars(c)$), where dom_divide_conquer is defined in Figure 10.

## 6.2 Set Bounds Propagation

As domain propagation may be prohibitively expensive to enforce for some problems, it is useful to investigate less strict notions of consistency. In such cases, we can speed propagation by simplifying the domains through approximation. Since set bounds under the subset partial ordering relation are one of the most commonly used approximations to a set domain, it seems natural to implement a set bounds propagator in the ROBDD framework. Only relatively minor changes are needed to the domain propagator to turn it into a set bounds propagator.

Given the ROBDD representation of a set domain, we can easily identify the corresponding set bounds. In an ROBDD-based set domain representation, the set bounds on a domain correspond to the *fixed* variables of the ROBDD representing the domain. We say an ROBDD variable $v$ is *fixed* if either for all nodes $n(v, t, e)$ $t$ is the constant 0 node, or for all nodes $n(v, t, e)$ $e$ is the constant 0 node, and such a node appears in every path from the root of the diagram to the "1" node.

Such nodes can be identified in a single pass over the domain ROBDD, in time proportional to its size. If $\phi$ is an ROBDD, we will write $[\![\phi]\!]$ to denote the ROBDD representing the conjunction of the fixed variables of $\phi$. If $\phi$ represents a set of sets $S$, then $[\![\phi]\!]$ represents $conv(S)$. An ROBDD $\phi$ where $\phi = [\![\phi]\!]$ is a stick ROBDD by definition.

**Example 6.1.** Let $\phi$ be the ROBDD depicted in Figure 1(c). Then $[\![\phi]\!]$ is the ROBDD of Figure 1(a).

Using this operation we can convert our domain propagator into a set bounds propagator by discarding all of the non-fixed variables from the domain ROBDDs after each propagation step. Suppose that $D(v)$ is a stick ROBDD for each $v \in \mathcal{V}$. If $c$ is a constraint, with $vars(c) = \{v_1, \ldots, v_n\}$, we define a set bounds propagator $sb(c)$ thus:

$$sb(c)(D)(v_i) = \left[\!\left[ \overline{\exists}_V(v_i)(B(c) \wedge \bigwedge_{j=1}^n D(v_j)) \right]\!\right] \tag{4}$$

Despite only relatively minor differences between the set bounds propagator and the domain propagator, the set bounds propagator is usually significantly faster than a domain propagator for two reasons. Firstly, as the domains $D(v)$ are all sticks, all of the ROBDD operations are cheap, compared to operations on the possibly very large ROBDDs representing arbitrary domains. The entire propagator can be implemented with $O(|B(c)|)$ complexity, since all of the other ROBDDs are sticks. Secondly, we can use the updated set bounds to simplify the propagator ROBDD $B(c)$. Since domains are monotonic decreasing in size, fixed variables will remain fixed up to backtracking, and so we can project them out of $B(c)$, thus reducing the size of the propagator ROBDD in future propagation steps. This leads us to the following implementation of the propagator:

$$\begin{aligned} \phi_0 &= B(c) \\ \phi_j &= \exists_{VAR(D(v_j))} D(v_j) \wedge \phi_{j-1} \quad 1 \le j \le n \\ sb(c)(D)(v_i) &= D(v_i) \wedge \overline{\exists}_{V(v_i)} [\![\phi_n]\!] \qquad 1 \le i \le n \end{aligned} \tag{5}$$

After this propagation step we can replace the representation of the constraint $B(c)$ by $\phi_n$ since the fixed variables will no longer have any new impact.

**Example 6.2.** Consider bounds propagation for the constraint $c \equiv v \subseteq w$ where $N = 3$. The ROBDD representation $B(c)$ is given in Figure 4. Assume the domains of $v$ and $w$ are respectively $[\{1\}, \{1, 2, 3\}]$, represented by the formula $v_1$, and $[\emptyset, \{1, 2\}]$, represented by the formula $\neg w_3$. The ROBDD $\phi_n \equiv \exists v_1 \exists w_3 \ B(c) \wedge v_1 \wedge \neg w_3$ is shown in Figure 11(a). We have that $[\![\phi_n]\!] = w_1 \wedge \neg v_3$. We can project out the fixed variables $v_1, w_1, v_3, w_3$ from $B(c)$ to get a new simplified form of the constraint $v_2 \to w_2$ shown in Figure 11(b).

This set bounds solver retains all of the modeling advantages of the domain solver, including the ability to easily conjoin and existentially quantify constraints, to remove intermediate variables and to form global constraints. In some cases this permits a substantial performance improvement over more traditional set bounds solvers.

Experimentally it appears that a direct implementation of Equation (4), written to use a divide and conquer approach to calculate the and-abstractions, is faster than an implementation of Equation (5), even if a divide and conquer approach is used in calculating the existential quantification. The former approach calculates fewer intermediate results, which leads to a faster propagator overall. Experimental results for the bounds propagator are given in Section 7.

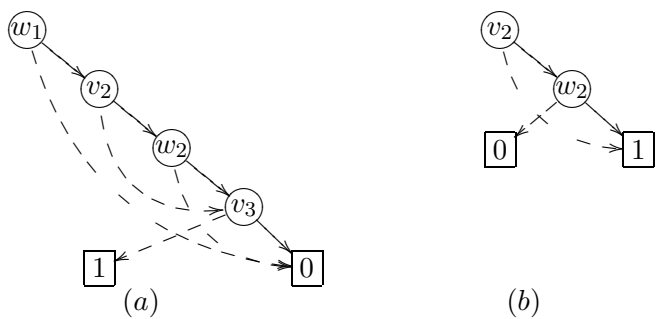

Figure 11: Set bounds propagation on the constraint $v \subseteq w$ showing (a) resulting ROBDD after conjunction with domains, and (b) simplified constraint ROBDD after removing fixed variables.

## 6.3 Split Domain Propagation

We can combine the set bounds propagator with the domain propagator to produce a more space efficient split domain propagator. By separating the domain representation into fixed and unfixed parts, we can reduce the total size of the representation, also hopefully speeding propagation.

One of the unfortunate characteristics of ROBDDs is that the size of the ROBDD representing a domain is highly dependent on the variable ordering. Consider an ROBDD representing a set domain which contains several fixed variables. If these variables do not appear at the beginning of the variable ordering, then the ROBDD representing the domain will in effect contain several copies of the sticks representing the fixed variables. For example, Figure 1(c) contains several copies of the stick in Figure 1(a). Since many of our ROBDD operations take time proportional to the product of the number of ROBDD nodes of their arguments, this overly large representation has a performance cost. We can solve this problem in two ways—either by reordering the ROBDD variables or by splitting up the domain representation.

Variable reordering is capable of eliminating redundancy in the representation of any individual domain, but in general cannot eliminate redundancy across a set of domains. By reordering the ROBDD variables, we can reduce the size of a domain by placing the fixed variables at the beginning of the variable order, thus removing the unnecessary duplication for that domain. Unfortunately, the variable order is a global property of all ROBDDs in existence, whereas the fixed variables of a domain are a local property specific to a particular domain, so there may not be a variable ordering that is optimal for all of the domains in a problem.

In the context of applying ROBDDs to the groundness analysis of logic programs, Bagnara (1996) demonstrated that the performance of an ROBDD-based program analyzer could be improved by splitting up ROBDDs into their fixed and non-fixed parts. We can apply the same technique here.

We split the ROBDD representing a domain $D(v)$ into a pair of ROBDDs $(L\overline{U}, R)$. $L\overline{U}$ is a stick ROBDD representing the lower and upper set bounds on $D(v)$, and $R$ is a remainder ROBDD representing the information on the unfixed part of the domain. Logically $D = L\overline{U} \wedge R$. We will write $L\overline{U}(D(v))$ and $R(D(v))$ to denote the $L\overline{U}$ and $R$ parts of $D(v)$ respectively.

The following results provide an upper bound of the size of the split domain representation:

**Lemma 6.1.** *Let $G$ be an ROBDD, and let $v$ be a fixed variable of $G$. Then $|\exists_v G| < |G|$.*

*Proof.* Since $v$ is a fixed variable, either for every node $n(v, t, f)$ in $G$ $t$ is the constant 0 node, or for every such node $f$ is the constant 0 node. Since a node $n(v, t, f)$ corresponds to the proposition $(v \wedge t) \vee (\neg v \wedge f)$, it is clear that $\exists_v n(v, t, f)$ corresponds to simply $t \vee f$, and moreover since $v$ is a fixed node one of $t$ or $f$ is zero. Hence the existential quantification of a fixed variable $v$ simply removes all nodes labelled $v$ from $D$. Since there is at least one such node, the result follows. $\qquad\square$

**Lemma 6.2.** *Let $G$ be an ROBDD, $L\overline{U} = [\![G]\!]$, and $R = \exists_{VAR(L\overline{U})} G$. Then $G \leftrightarrow L\overline{U} \wedge R$ and $|L\overline{U}| + |R| \leq |G|$.*

*Proof.* The result $G \leftrightarrow L\overline{U} \wedge R$ is straightforward, so we only prove the result on sizes.

Suppose that $VAR(L\overline{U}) = \{v_1, \ldots, v_n\}$ is the set of fixed variables of $G$. Then, since $|\exists_{v_1} G| < |G|$, $1 + |\exists_{v_1} G| \leq |G|$. By repeating this operation for each $v_i$, we obtain $n + |\exists_{VAR(L\overline{U})} G| \leq |G|$. But as $L\overline{U}$ is a stick, trivially $|L\overline{U}| = n$, and as $R = \exists_{VAR(L\overline{U})} G$ by definition this is the required inequality. $\qquad\square$

Note that $|D|$ can be $O(|L\overline{U}| \times |R|)$. For example, considering the ROBDDs in Figure 1 where $L\overline{U}$ is shown in (a), $R$ in (b), and $D = L\overline{U} \wedge R$ in (c), we have that $|L\overline{U}| = 5$ and $|R| = 9$ but $|D| = 9 + 4 \times 5 = 29$.

We now show how to construct a propagator on split domains. Firstly, we eliminate any fixed variables (as in the bounds propagator) and then apply domain propagation on the remainder domains $R$. The propagator produces a new pair $(L\overline{U}, R)$ consisting of new fixed variables and a new remainder. This process is shown below:

$$\phi_0 = B(c)$$
$$\phi_j = \exists_{VAR(L\overline{U}(D(v_j)))} (L\overline{U}(D(v_j)) \wedge \phi_{j-1})$$
$$\delta_i = \overline{\exists}_{V(v_i)} \left( \phi_n \wedge \bigwedge_{j=1}^{n} R(D(v_j)) \right) \tag{6}$$
$$\beta_i = L\overline{U}(D(v_i)) \wedge [\![\delta_i]\!]$$
$$dom(c)(D)(v_i) = (\beta_i, \exists_{VAR(\beta_i)} \delta_i)$$

For efficiency the $\delta_i$ components can be calculated using the divide-and-conquer approach described for the domain propagator.

The split domain representation has three main advantages. Proposition 6.2 tells us that the split domain representation is no larger than the original domain representation. However, often the split representation is substantially smaller, which can lead to improvements in propagation performance. The split solver can also use the propagator simplification technique from the bounds solver by abstracting the fixed variables out of the propagator ROBDDs. Finally using the split solver we can mix the usage of domain and bounds propagators in the same problem.

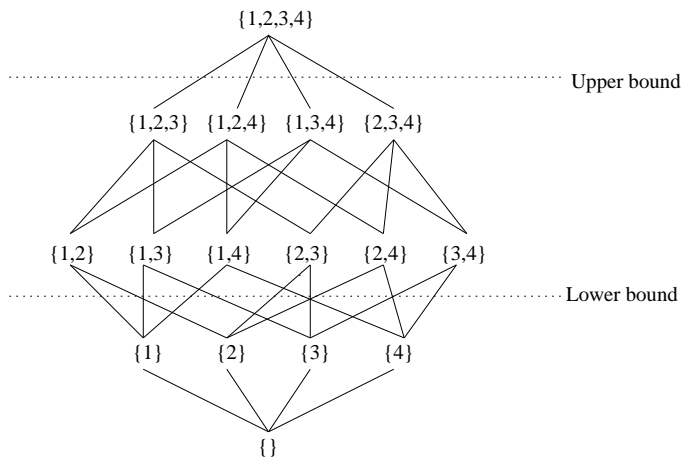

Figure 12: The set interval $[\emptyset, \{1, 2, 3, 4\}]$ with upper and lower cardinality bounds $u = 3$ and $l = 2$ respectively

Experimental results for the split domain propagator are given in Section 7.

## 6.4 Cardinality Bounds Propagation

Given that we are able to model a set bounds propagator using ROBDDs, it is also appropriate to consider how we might model other levels of consistency for set constraint problems. One level of consistency that is commonly used (Azevedo, 2002; Müller, 2001) is that of combined set bounds and cardinality consistency, in which upper and lower bounds on the cardinality of each domain are maintained in addition to the bounds under the subset partial ordering. This hybrid approach allows a more accurate representation of some domains, particularly those with constrained cardinality, which are common in set problems.

**Example 6.3.** Figure 12 depicts the set interval $[\emptyset, \{1, 2, 3, 4\}]$, together with lower and upper cardinality bounds 2 and 3 respectively. In general an interval consists of a large number of sets, making it a crude approximation to a set domain. Cardinality bounds permit a more fine grained representation by effectively allowing us to select a subset of the rows of the lattice diagram.

Just as we were able to implement a set bounds propagator using the domain propagator together with a function which extracts the set bounds of a domain, so too can we create a combined set bounds and set cardinality propagator. Here we will extend the split domain solver by simplifying the "remainder" component of a split domain to an ROBDD representing its cardinality bounds.

bdd_count_cardinality$(D, V)$:
    **if** $(D = 0)$ **return** $\langle \infty, -\infty \rangle$
    **else if** $(D = 1)$
        **if** $(|V| = 0)$ **return** $\langle 0, 0 \rangle$
        **else**
            $\langle v_1, v_2, \ldots, v_n \rangle := V$
            **return** $\langle 0, n \rangle$
    **else**
        **if** $(|V| = 0)$ **error**
        $n(v, t, e) := D$
        $\langle v_1, v_2, \ldots, v_n \rangle := V$
        **if** $(v_1 \succ v)$ **error**
        **else if** $(v = v_1)$
            $\langle l_t, u_t \rangle := $ bdd_count_cardinality$(t, \langle v_2, \ldots, v_n \rangle)$
            $\langle l_e, u_e \rangle := $ bdd_count_cardinality$(e, \langle v_2, \ldots, v_n \rangle)$
            **return** $\langle \min(l_t + 1, l_e), \max(u_t + 1, u_e) \rangle$
        **else**
            $\langle l, u \rangle := $ bdd_count_cardinality$(D, \langle v_2, \ldots, v_n \rangle)$
            **return** $\langle l, u + 1 \rangle$

bdd_card_bounds$(D, V)$:
    $\langle l, u \rangle := $ bdd_count_cardinality$(D, V)$
    **if** $(l = \infty$ **or** $u = -\infty)$ **return** 0
    **return** $card(V, l, u)$

Figure 13: An algorithm to determine the cardinality bounds on the domain of a set variable represented by an ROBDD $D$, where $V$ is the vector of bits in the Boolean representation of the set variable

As before, we need a method for extracting an ROBDD representing the cardinality bounds of an arbitrary domain ROBDD. We perform this operation in two stages. Firstly we define a function bdd_count_cardinality which takes an ROBDD representing a set domain and returns upper and lower bounds on its cardinality. We can represent these bounds in ROBDD form by constructing a new cardinality constraint ROBDD as described in Section 3.2.

An implementation of bdd_count_cardinality is shown in Figure 13. This function can be implemented to run in $O(|D||V|)$ time if dynamic programming/caching is used to save the results of the intermediate recursive calls. In practice since $D$ and $V$ are highly interrelated, it is $O(|D|)$. In our implementation we utilise the global cache mechanism of the ROBDD library, which also permits caching of partial results between multiple calls to bdd_count_cardinality.

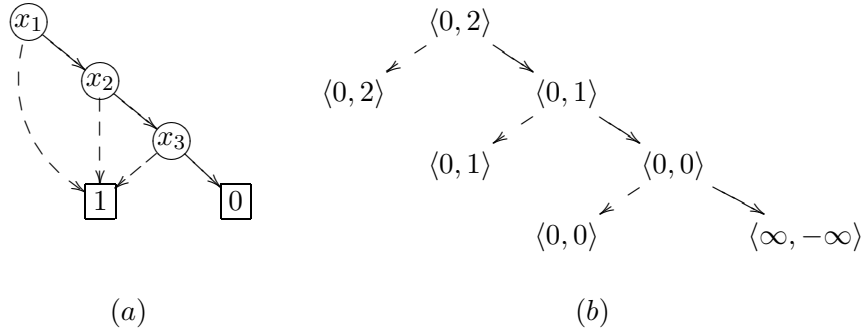

Figure 14: Cardinality propagation example showing (a) resulting ROBDD projected onto $x$, and (b) calculation of cardinality bounds

We can now use this function to construct a function `bdd_card_bounds` which takes an ROBDD $D$ and a list of Boolean variables $V$ and returns a new ROBDD representing the bounds on the cardinality of the solutions of $D$. The implementation of `bdd_card_bounds` is shown in Figure 13.

The algorithm for a split set bounds and set cardinality propagator for a constraint $c$ is given by $sbc(c)$ in the following equation:

$$
\begin{aligned}
\phi_0 &= B(c) \\
\phi_j &= \exists_{VAR(L\overline{U}(D(v_j)))}(L\overline{U}(D(v_j)) \wedge \phi_{j-1}) \\
\delta_i &= \overline{\exists}_{V(v_i)}\left(\phi_n \wedge \bigwedge_{j=1}^{n} R(D(v_j))\right) \\
\beta_i &= L\overline{U}(D(v_i)) \wedge [\![\delta_i]\!] \\
sbc(c)(D)(v_i) &= (\beta_i, \mathsf{bdd\_card\_bounds}(\exists_{VAR(\beta_i)}\delta_i, V(v_i) \setminus VAR(\beta_i)))
\end{aligned}
\tag{7}
$$

Note that we only keep the cardinality bounds on the remaining non-fixed Boolean variables for a set variable $v$ rather than on all the original variables $V(v)$, since we do not need to consider fixed variables again, and it leads to a slightly smaller cardinality ROBDD. As usual Equation 7 should be implemented using a divide-and-conquer approach for efficiency. Experimental results for this propagator are shown in Section 7.

**Example 6.4.** We illustrate set bounds and cardinality propagation on the constraint $c \equiv lexlt(x, y)$ whose ROBDD $B(c)$ is shown in Figure 7(b). Assume the original domains for $x$ and $y$ are universal, so $D(x) = D(y) = [\emptyset, \{1, 2, 3\}]$, represented by the ROBDD 1. The ROBDD for $\delta_x = \overline{\exists}_{V(x)}B(c)$ is shown in Figure 14(a), and $\beta_x = [\![\delta_x]\!] = 1$. The tree of calculations for $\mathsf{bdd\_card\_bounds}(\delta_x, \langle x_1, x_2, x_3 \rangle)$ is shown in Figure 14(b). Overall the cardinality of $x$ is determined to be in the range $[0, 2]$.

### 6.5 Lexicographic Bounds Propagation

An alternative form of set consistency proposed by Sadler and Gervet (2004) is to maintain bounds under a lexicographic ordering in addition to set bounds. The lexicographic ordering is a total ordering on sets which embeds the subset partial ordering. Bounds under the lexicographic ordering alone are not sufficient to express the effects of many constraints (in particular the inclusion of a single element), so Sadler and Gervet constructed a hybrid solver which combines lexicographic bounds with traditional set bounds. They demonstrated that a set solver based upon lexicographic bounds consistency techniques produced stronger propagation than a traditional set bounds solver, although this came at a substantial cost in propagation performance. However, given that the use of the ROBDD representation leads to a performance improvement in the case of set bounds propagation, it is worth investigating the performance of an ROBDD-based lexicographic bounds propagator.

The lexicographic bounds of a domain can be very compactly represented as an ROBDD. Like set bounds, an ROBDD representing the upper or lower lexicographic bounds of a domain is an ROBDD of size $O(N)$, and so is the combination. Because these ROBDDs are compact this hopefully leads to fast propagation. Moreover, given an ROBDD domain, it is very easy to extract the lexicographic bounds on that domain in a single pass. By a process analogous to the construction of the bounds and cardinality propagator, we can use a split domain propagator combined with a function which determines the lexicographic bounds of a domain to construct a highly efficient lexicographic bounds propagator.

We define two functions `bdd_lex_lower_bound` and `bdd_lex_upper_bound`. These functions, given an ROBDD representing the domain $D$ of a variable $V$ together with a list of Boolean variables $B$ corresponding to the bits of $V$, return the lower and upper lexicographic bounds (respectively) of $D$. `bdd_lex_lower_bound` is implemented as shown in Figure 15 (the implementation of `bdd_lex_upper_bound` is very similar).

We define:

$$\mathsf{bdd\_lex\_bounds}(D, B) = \mathsf{bdd\_lex\_lower\_bound}(D, B) \wedge \mathsf{bdd\_lex\_upper\_bound}(D, B)$$

The split set and lexicographic bounds propagator can then be implemented exactly as in Equation (7), using `bdd_lex_bounds` in place of `bdd_card_bounds`.

**Example 6.5.** Consider lexicographic bounds propagation on the constraint $c \equiv |s| = 2$ where $s \subseteq \{1, 2, 3, 4, 5\}$. The ROBDD $B(c)$ is shown in Figure 3(b) and initial domain $D(s) = [\emptyset, \{1, 2, 3, 4, 5\}]$. Then $\delta_s = B(c)$ and $\beta_s = 1$ so the call to `bdd_lex_bounds` calculates the lower bounds lexicographic ROBDD shown in Figure 16(a), the upper bounds lexicographic ROBDD shown in Figure 16(b), and final answer the conjunction shown in Figure 16(c). Note that we have lost some information relative to the original cardinality ROBDD.

As observed in Section 5.3, the lexicographic ordering for set variables actually corresponds to a numeric ordering on integer variables, so a pure lexicographic bounds propagator would also be coincidentally an integer bounds propagator.

Experimental results for the lexicographic bounds propagator are given in Section 7.

```
bdd_lex_lower_bound(D, B):
    if (|B| = 0 or D = 1) return 1
    if (D = 0) error
    n(v, t, e) := D
    ⟨b₁, ..., bₙ⟩ := B
    if (b₁ ≻ v) error
    else if (b₁ = v and e = 0)
        return b₁ ∧ bdd_lex_lower_bound(t, ⟨b₂, ..., bₙ⟩)
    else
        if (b₁ = v) r := bdd_lex_lower_bound(e, ⟨b₂, ..., bₙ⟩)
        else r := bdd_lex_lower_bound(D, ⟨b₂, ..., bₙ⟩)
        return b₁ ∨ (¬b₁ ∧ r)
```

Figure 15: An algorithm to extract the lower lexicographic bounds on the domain of a set variable represented by an ROBDD $D$, where $B$ is the vector of (non-fixed) bits in the Boolean representation of the set variable

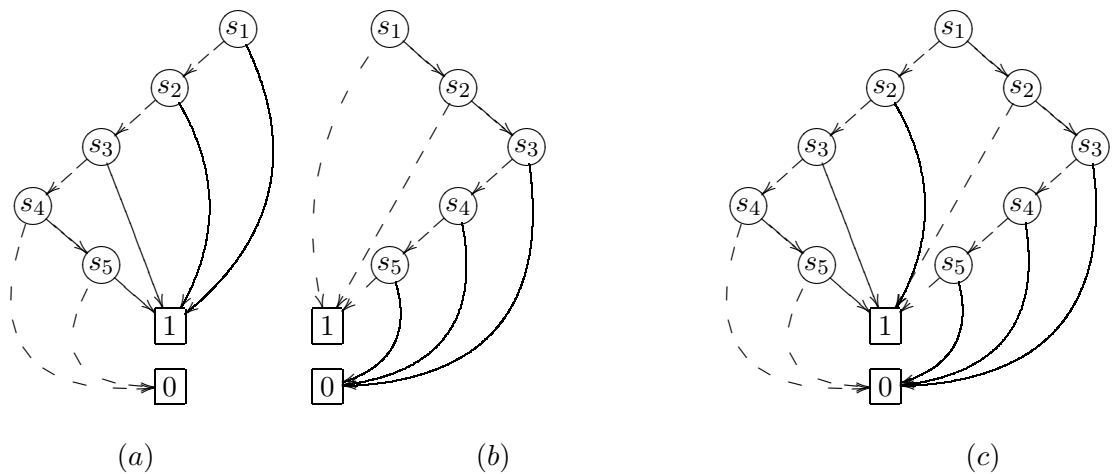

$(a)$ $(b)$ $(c)$

Figure 16: Calculating lexicographic (a) lower and (b) upper bounds and (c) their conjunction.

## 7. Experiments

We have implemented a set solver using the ideas described in this paper. Our implementation is written primarily in MERCURY (Somogyi, Henderson, & Conway, 1996), using an interface to the C language ROBDD library CUDD as a platform for ROBDD manipulations (Somenzi, 2004). The ROBDD library is effectively treated as a black box.

We used our solver to implement a series of standard constraint benchmarks as described below. Many of these problems are in the CSPLib library of constraint satisfaction problems (Gent, Walsh, & Selman, 2004). For the purposes of comparison, we also implemented our benchmarks using the ic_sets library of ECL$^i$PS$^e$ v5.6[2] and the finite sets library of Mozart v1.3.0. We conducted all of our experiments on a cluster of 8 identical 2.8GHz Pentium 4 machines, each with 1 Gb of RAM and running Debian GNU/Linux 3.1. All three solvers were limited to 1 Gb of memory to minimise swapping. All experiments were repeated three times, and the lowest time out of the three runs was taken as the result.

## 7.1 Steiner Systems

A commonly used benchmark for set constraint solvers is the calculation of small Steiner systems. A Steiner system $S(t, k, N)$ is a set $X$ of cardinality $N$ and a collection $C$ of subsets of $X$ of cardinality $k$ (called "blocks"), such that any $t$ elements of $X$ are in exactly one block. Steiner systems have been extensively studied in combinatorial mathematics. If $t = 2$ and $k = 3$, we have the so-called Steiner triple systems, which are often used as benchmarks (Gervet, 1997; Azevedo, 2002; Müller, 2001). Any Steiner system must have exactly $m = \binom{N}{t} / \binom{k}{t}$ blocks (Theorem 19.2 of van Lint & Wilson, 2001).

It is natural to model a Steiner system using set variables $s_1, \ldots, s_m$, where each set variable corresponds to a single block, subject to the following constraints:

$$\bigwedge_{i=1}^{m} (|s_i| = k) \tag{8}$$

$$\wedge \bigwedge_{i=1}^{m-1} \bigwedge_{j=i+1}^{m} (\exists u_{ij} \; u_{ij} = s_i \cap s_j \wedge |u_{ij}| \leq (t-1)) \wedge (s_i < s_j) \tag{9}$$

The lexicographic ordering constraint $s_i < s_j$ has been added to remove symmetries from the problem formed by permuting the blocks.

A necessary condition for the existence of a Steiner system is that $\binom{N-i}{t-i} / \binom{k-i}{t-i}$ is an integer for all $i \in \{0, 1, \ldots, t-1\}$ (van Lint & Wilson, 2001); we say a set of parameters $(t, k, N)$ is admissible if it satisfies this condition. In order to choose test cases, we ran each solver on every admissible set of $(t, k, N)$ values for $N < 32$. Results are shown for every test case that at least one solver was able to solve within a time limit of 10 minutes. "×" denotes abnormal termination due to exceeding an arbitrary limit on the maximum number of ROBDD variables imposed by the CUDD package, while "—" denotes failure to complete a testcase within the time limit.

In all cases we use a sequential variable-ordering heuristic, and a value-ordering heuristic that chooses the largest unfixed value within a variable's domain. Once a variable and an unfixed value have been chosen for labeling, either that value is a member of the set represented by the variable or it is not. The order in which we try the two alternatives has a significant effect on the performance of the solver. In this case we elect to choose the element-not-in-set option first.

---

2. Very recently a new sets library Cardinal has been added to ECL$^i$PS$^e$ which supports better cardinality reasoning. Unfortunately we cannot directly compare with it on these benchmarks since it does not

Table 2: Results for Steiner systems, with split constraints and intermediate variables.

| Testcase | ECL$^i$PS$^e$ | | MOZART | | Bounds | | Domain | | LU+R | | LU+Lex | | LU+Card | |
|---|---|---|---|---|---|---|---|---|---|---|---|---|---|---|
| | Time/s | Fails | Time/s | Fails | Time/s | Fails | Time/s | Fails | Time/s | Fails | Time/s | Fails | Time/s | Fails |
| S(2,3,7) | 0.3 | 10 | 0.1 | 21 | **<0.1** | 10 | 0.1 | **0** | 0.1 | **0** | 0.1 | 4 | **<0.1** | 2 |
| S(3,4,8) | 0.5 | 21 | **0.1** | 52 | **0.1** | 21 | 0.4 | **0** | 0.4 | **0** | 0.4 | 4 | **0.1** | 4 |
| S(2,3,9) | 7.7 | 1394 | **1.0** | 5102 | 1.6 | 1394 | **1.0** | **100** | 1.3 | **100** | 3.3 | 421 | 2.4 | 1072 |
| S(2,4,13) | 1.8 | 313 | **0.4** | 1685 | 0.6 | 313 | 1.7 | **32** | 1.5 | **32** | 2.1 | 127 | 0.7 | 157 |
| S(2,3,15) | 3.6 | 65 | **0.5** | 354 | 2.2 | 65 | 20.2 | 0 | 19.6 | 0 | 20.4 | 127 | 3.3 | 41 |
| S(3,4,16) | **67.5** | **289** | — | | × | × | × | × | × | × | × | × | × | × |
| S(2,5,21) | 3.2 | 421 | **0.4** | 668 | 2.3 | 421 | 110.2 | 0 | 59.8 | 0 | 21.5 | 139 | 2.6 | 124 |
| S(3,6,22) | **49.7** | **1619** | — | | × | × | × | × | × | × | × | × | × | × |

Table 3: Results for Steiner systems, with merged constraints and no intermediate variables.

| Testcase | Bounds | | Domain | | LU+R | | LU+Lex | | LU+Card | |
|---|---|---|---|---|---|---|---|---|---|---|
| | Time/s | Fails | Time/s | Fails | Time/s | Fails | Time/s | Fails | Time/s | Fails |
| S(2,3,7) | **<0.1** | 8 | **<0.1** | **0** | **<0.1** | **0** | **<0.1** | **0** | **<0.1** | **0** |
| S(3,4,8) | **0.1** | 18 | **0.1** | **0** | **0.1** | **0** | **0.1** | **0** | **0.1** | **0** |
| S(2,3,9) | 0.2 | 325 | **0.1** | **9** | **0.1** | **9** | **0.1** | 11 | 0.2 | 113 |
| S(2,3,13) | — | — | **109.2** | **24723** | 144.6 | **24723** | 518.1 | 30338 | — | — |
| S(2,4,13) | **0.1** | 157 | **0.1** | **0** | **0.1** | **0** | 0.4 | 11 | **0.1** | 27 |
| S(2,3,15) | **0.4** | 56 | 1.3 | **0** | 1.4 | **0** | 2.9 | **0** | 0.7 | 32 |
| S(2,4,16) | 421.4 | 522706 | **0.6** | **15** | **0.6** | **15** | 2.5 | 16 | 577.0 | 209799 |
| S(2,6,16) | — | — | **80.7** | **15205** | 82.7 | **15205** | — | — | — | — |
| S(3,4,16) | **9.7** | 274 | 548.7 | **0** | 485.3 | **0** | 428.9 | **0** | 18.4 | 162 |
| S(2,5,21) | **0.5** | 413 | 1.4 | **0** | 1.4 | **0** | 32.9 | **0** | 0.6 | 116 |
| S(3,6,22) | **8.3** | 1608 | — | — | — | — | — | — | 12.7 | **381** |
| S(2,3,31) | **23.3** | 280 | — | — | — | — | — | — | 48.6 | **224** |

Table 4: All-solutions results on Steiner systems. "—" denotes failure to complete a test case within one hour

| Problem | Solns. | ECL$^i$PS$^e$ | | Bounds | | Domain | | LU+R | | LU+Lex | | LU+Card | |
|---|---|---|---|---|---|---|---|---|---|---|---|---|---|
| | | time/s | fails | time/s | fails | time/s | fails | time/s | fails | time/s | fails | time/s | fails |
| S(2,3,7) | 30 | 16.3 | 3,015 | 0.2 | 537 | **0.1** | **47** | **0.1** | **47** | 0.2 | 76 | 0.3 | 267 |
| S(3,4,8) | 30 | — | — | 726.5 | 610271 | **1.4** | **492** | 2.1 | **492** | 22.4 | 3248 | 951.2 | 431801 |
| S(2,3,9) | 840 | — | — | 398.1 | 391691 | **23.4** | **16794** | 37.8 | **16794** | 110.4 | 29133 | 593.3 | 224131 |
| S(2,6,16) | 0 | — | — | — | — | **80.9** | **15205** | 83.0 | **15205** | — | — | — | — |

In order to compare the raw performance of the various solvers, irrespective of any modeling advantages of the ROBDD-based solver, we performed experiments using a model of the problem which contains only primitive constraints and makes use of intermediate variables. This "split" model contains $m$ unary constraints corresponding to Equation (8) and $3m(m-1)/2$ binary constraints corresponding to Equation (9) (containing intermediate variables $u_{ij}$). The same model was used in the ECL$^i$PS$^e$, Mozart and ROBDD-based solvers, permitting a direct comparison of propagation performance. Results for this model are shown in Table 2. In particular, observe that with this model ECL$^i$PS$^e$ and the ROBDD-based bounds solver produce the same number of failures, demonstrating that the search spaces explored by the two solvers are identical.

One of the main strengths of the ROBDD-based modeling approach is that it gives us the freedom to merge arbitrary constraints and existentially quantify away intermediate variables, allowing us to model set constraint problems more efficiently. In the case of Steiner systems, this allows us to model the problem as $m$ unary constraints corresponding to Equation (8), and just $m(m-1)/2$ binary constraints $\psi_{ij}$ of the form $\psi_{ij} = (|s_i \cap s_j| \leq (t-1)) \wedge (s_i < s_j)$. The binary constraints do not contain the intermediate variables $u_{ij}$—they are not required since they can be existentially quantified out of the ROBDD representation. Experimental results for this revised model are shown in Table 3.

In all cases the revised model propagates much more strongly than the original model, leading to a substantial decrease in solution time. In addition, the decrease in the number of set variables required permits the solution of larger test cases. Clearly it is beneficial to remove intermediate variables and merge constraints.

Despite weaker propagation the ROBDD bounds solver is often the fastest method of finding a single solution to a Steiner System. In order to determine whether this was due to the efficiency of the solver, or whether the solver was just "lucky" in finding a solution quickly, we also ran experiments to find all solutions of the Steiner systems up to reordering of the blocks. The results for all test cases that at least one of the solvers was able to solve within a time limit of one hour are shown in Table 4. In all cases the reduction both of time and number of fails demonstrate the superiority of the propagation approaches based on domain consistency.

## 7.2 Social Golfers

Another problem often used as a set benchmark is the "Social Golfers" problem (problem `prob010` of CSPLib). The aim of this problem is to arrange $N = g \times s$ golfers into $g$ groups of $s$ players for each of $w$ weeks, such that no two players play together more than once. We can model this problem as a set constraint problem using a $w \times g$ matrix of set variables $v_{ij}$, where $1 \leq i \leq w$ is the week index and $1 \leq j \leq g$ is the group index.

---

support lexicographic ordering constraints. Testing without lexicographic orderings showed that it was about 2–3 times slower than LU + Card.

We use the following model of the problem:

$$\bigwedge_{i=1}^{w}(\texttt{partition}^<(v_{i1},\ldots,v_{ig})) \wedge \bigwedge_{i=1}^{w}\bigwedge_{j=1}^{g}|v_{ij}| = s$$

$$\wedge \bigwedge_{\substack{i,j\in\{1,\ldots,w\}\\i\neq j}}\bigwedge_{k,l\in\{1,\ldots,g\}}|v_{ik}\cap v_{jl}| \leq 1 \wedge \bigwedge_{i=1}^{w-1}\bigwedge_{j=i+1}^{w}v_{i1} \leq v_{j1} \tag{10}$$

The $\texttt{partition}^<$ global constraint is a combined partitioning and lexicographical ordering constraint, formed by merging the $\texttt{partition}$ constraint of Section 4.2 with constraints imposing a lexicographic order on the variables. This constraint is trivial to construct using ROBDDs, but is not available in either ECL$^i$PS$^e$ or Mozart.[3]

Results for each of the solvers are shown in Table 5 and Table 6. In the former table, a sequential smallest-element-in-set labeling strategy was used to enable a fair comparison of propagation performance, whereas in the latter table a first-fail labeling strategy was used in order to give a measure of the peak performance of each solver. For every test case in both tables, with a single exception (5-8-3), at least one of the ROBDD-based solvers performs equal or better to both ECL$^i$PS$^e$ and Mozart. It should also be observed that when using a first-fail labeling strategy, the domain and split domain solvers are the only solvers able to solve every test case.

There are several other features of the results that are worth noting. As in the case of Steiner systems, the ROBDD-based set bounds solver is often the fastest, despite weak propagation. Amongst the solvers with stronger propagation, the split domain solver is almost always faster than the original domain solver due to smaller domain sizes. It is, however, slower than the original domain solver in the presence of backtracking (due to the requirement to trail more values—in particular the propagator ROBDDs). The lexicographic bounds solver is almost as effective as the domain solvers in restricting search space, although it is usually outperformed by the domain and bounds solvers.

## 7.3 Weighted Hamming Codes

The problem of finding maximal Hamming Codes can be modelled as a set constraint problem.

We define an $l$-bit codeword to be a bit-string (or vector of Boolean values) of length $l$. Given two $l$-bit codewords $A$ and $B$, we define the *Hamming distance* $d(A, B)$ between $A$ and $B$ to be the number of positions at which the two bit-strings differ. An $(l, d)$-*Hamming Code* is a set of $l$-bit codewords such that the Hamming distance between any two codewords in the set is at least $d$.

Given a codeword length $l$ and the minimum Hamming distance $d$, the problem is to construct a Hamming code with the largest possible number of codewords. A variant of this problem, used as a benchmark by Sadler and Gervet (2004), has the additional requirement that each codeword have a fixed weight $w$, where the weight of a codeword is defined to be the number of "1" bits that codeword contains. We will denote an instance of this problem by $H(l, d, w)$.

---

3. It would, of course, be possible to implement such a constraint in both ECL$^i$PS$^e$ and Mozart, but such an implementation would be a fairly laborious process. A strength of the ROBDD-based modeling approach is that we can construct global constraints with no extra code.

Table 5: First-solution performance results on the Social Golfers problem, using a sequential "smallest-element-in-set" labeling strategy. Time and number of failures are given for all solvers. "—" denotes failure to complete a test case within 10 minutes. The cases 5-4-3, 6-4-3, and 7-5-5 have no solutions

| Problem | ECL$^i$PS$^e$ | | MOZART | | Bounds | | Domain | | LU+R | LU+Lex | | LU+Card | |
|---|---|---|---|---|---|---|---|---|---|---|---|---|---|
| $w$-$g$-$s$ | time /s | fails | time /s | fails | time /s | fails | time /s | fails | time /s | time /s | fails | time /s | fails |
| 2-5-4 | 7.6 | 10468 | 1.0 | 7638 | **0.1** | 30 | **0.1** | **0** | **0.1** | **0.1** | 3 | **0.1** | 5 |
| 2-6-4 | 49.2 | 64308 | 6.4 | 42346 | 0.6 | 2036 | 0.2 | **0** | **0.1** | 0.7 | 194 | 0.4 | 326 |
| 2-7-4 | 95.1 | 114818 | 10.9 | 66637 | 1.7 | 4447 | **0.4** | **0** | **0.4** | 2.3 | 692 | 1.6 | 1608 |
| 2-8-5 | — | — | — | — | — | — | 2.0 | **0** | **1.6** | — | — | — | — |
| 3-5-4 | 12.5 | 14092 | 2.5 | 10311 | **0.1** | 30 | 0.3 | **0** | 0.3 | 0.4 | 3 | 0.2 | 5 |
| 3-6-4 | 76.3 | 83815 | 14.0 | 51134 | 1.6 | 2039 | 1.6 | **0** | **1.4** | 2.2 | 194 | 1.9 | 328 |
| 3-7-4 | 146.8 | 146419 | 27.3 | 88394 | **4.6** | 4492 | 8.9 | **0** | 8.4 | 7.6 | 695 | 5.1 | 1629 |
| 4-5-4 | 14.1 | 14369 | 3.9 | 10715 | **0.2** | 30 | 0.8 | **0** | 0.6 | 0.7 | 3 | 0.4 | 5 |
| 4-6-5 | — | — | — | — | 21.9 | 12747 | 118.6 | **0** | 80.7 | **19.3** | 499 | 32.0 | 2122 |
| 4-7-4 | 169.3 | 149767 | 46.4 | 90712 | **8.7** | 4498 | — | — | 481.6 | 14.1 | **696** | 10.5 | 1632 |
| 4-9-4 | 27.3 | 19065 | 7.8 | 12489 | **2.6** | 71 | — | — | — | 22.2 | **8** | 8.9 | 33 |
| 5-3-4 | — | — | — | — | 113.9 | 63642 | **28.6** | **5165** | 32.1 | 88.9 | 10210 | 202.7 | 50542 |
| 5-5-4 | 350.6 | 199632 | 217.7 | 416889 | 7.0 | 2686 | 3.8 | **41** | **2.3** | 12 | 313 | 20.8 | 1584 |
| 5-7-4 | — | — | — | — | **14.6** | 4583 | — | — | — | 23.7 | **700** | 17.5 | 1683 |
| 5-8-3 | 5.0 | 2229 | **0.9** | 1820 | 1.1 | 14 | 9.2 | **0** | 7.8 | 3.2 | 3 | 2.1 | 4 |
| 6-4-3 | — | — | — | — | 158.2 | 61770 | **20.3** | **2132** | 23.0 | 60.8 | 4506 | 293.5 | 49966 |
| 6-5-3 | 458.9 | 240296 | 287.4 | 471485 | 4.1 | 1455 | 1.8 | **82** | **1.5** | 4.1 | 202 | 8.3 | 1078 |
| 6-6-3 | 3.3 | 1462 | 1.0 | 1462 | **0.5** | 5 | 1.8 | **0** | 1.4 | 1.2 | 0 | 0.7 | 1 |
| 7-5-5 | — | — | — | — | — | — | 0.5 | **0** | **0.4** | 2.4 | 0 | 1.3 | 22 |

As proposed by Müller and Müller (1997), we can model this problem for $n$ codewords using $n$ set variables $S_i$, where $1 \leq i \leq n$. A codeword $C_i$ corresponds to the characteristic function of the set $S_i$, i.e. bit $j$ is set in codeword $C_i$ if and only if $j \in S_i$. The Hamming distance $d(C_i, C_j)$ between two codewords $C_i$ and $C_j$ can be calculated from the associated sets $S_i$ and $S_j$ thus:

$$d(C_i, C_j) = l - |S_i \cap S_j| - |\{1, \ldots, l\} \setminus (S_i \cup S_j)|$$

We can remove symmetries created by permuting the codewords by introducing lexicographic ordering constraints $S_i < S_j$, for all $1 \leq i < j \leq n$. The complete model of the

Table 6: First-solution performance results on the Social Golfers problem, using a first-fail "smallest-element-in-set" labeling strategy. Time and number of failures are given for all solvers. The cases 5-4-3, 6-4-3, and 7-5-5 have no solutions

| Problem | ECL$^i$PS$^e$ | | MOZART | | Bounds | | Domain | | LU+R | LU+Lex | | LU+Card | |
|---|---|---|---|---|---|---|---|---|---|---|---|---|---|
| | time /s | fails | time /s | fails | time /s | fails | time /s | fails | time /s | time /s | fails | time /s | fails |
| *w-g-s* | | | | | | | | | | | | | |
| 2-5-4 | 7.9 | 10468 | 1.1 | 7638 | **0.1** | 30 | 0.1 | 0 | **0.1** | 0.1 | 3 | 0.4 | 447 |
| 2-6-4 | 51.3 | 64308 | 6.5 | 42346 | 0.6 | 2036 | 0.2 | 0 | **0.1** | 0.7 | 186 | 3.8 | 3820 |
| 2-7-4 | 99.9 | 114818 | 11.1 | 66637 | 1.7 | 4447 | **0.4** | 0 | **0.4** | 2.0 | 390 | 4.6 | 6424 |
| 2-8-5 | — | — | — | — | — | — | 1.9 | 0 | **1.6** | — | — | — | — |
| 3-5-4 | 14.5 | 14092 | 2.6 | 10311 | **0.1** | 44 | 0.3 | 0 | 0.3 | 0.4 | 5 | 1.3 | 481 |
| 3-6-4 | 91.8 | 83815 | 15.1 | 51134 | 1.9 | 2361 | 1.1 | 0 | **0.9** | 2.4 | 209 | 13.7 | 3722 |
| 3-7-4 | 183.0 | 146419 | 28.7 | 88394 | 5.2 | 5140 | 1.9 | 0 | **1.7** | 6.2 | 512 | 15.6 | 6067 |
| 4-5-4 | 18.0 | 14369 | 4.1 | 10715 | **0.3** | 47 | 0.8 | 0 | 0.6 | 0.7 | 7 | 2.6 | 494 |
| 4-6-5 | — | — | — | — | **38.5** | 19376 | 62.5 | 0 | 40.1 | 24.6 | 607 | — | — |
| 4-7-4 | 243.9 | 149767 | 49.6 | 90712 | 9.9 | 5149 | 6.5 | 0 | **5.1** | 10.6 | 405 | 29.5 | 5717 |
| 4-9-4 | 40.9 | 19065 | 8.3 | 12489 | **2.7** | 143 | 152.0 | 0 | 107.4 | 22.6 | 13 | 12.5 | 516 |
| 5-4-3 | — | — | — | — | 187.5 | 103972 | **23.2** | 3812 | 26.0 | 91.4 | 10422 | 309.1 | 72669 |
| 5-5-4 | 394.9 | 199632 | 224.5 | 416889 | 4.5 | 2388 | 2.5 | 18 | **1.7** | 23.4 | 776 | 41.4 | 4730 |
| 5-7-4 | — | — | — | — | 17.6 | 5494 | 18.2 | 0 | **12.8** | 16.9 | 447 | 47.1 | 5473 |
| 5-8-3 | 5.4 | 2229 | **0.9** | 1820 | 1.1 | 19 | 4.5 | 0 | 3.9 | 3.4 | 2 | 2.7 | 83 |
| 6-4-3 | — | — | — | — | 234.2 | 90428 | **14.6** | 1504 | 15.2 | 54.0 | 4013 | 349.9 | 59805 |
| 6-5-3 | 501.0 | 240296 | 294.2 | 471485 | 1.6 | 495 | 1.2 | 34 | **1.0** | 58.2 | 3787 | 473.6 | 68673 |
| 6-6-3 | 3.6 | 1462 | **1.0** | 1462 | — | — | 1.6 | **7** | 1.3 | 5.1 | 292 | **1.0** | 8 |
| 7-5-3 | — | — | — | — | — | — | 16.9 | **528** | **13.1** | 288.8 | 9829 | | |
| 7-5-5 | — | — | — | — | — | — | 0.5 | 0 | **0.4** | 2.4 | 0 | 1.3 | 22 |

problem is:

$$\bigwedge_{i=1}^{n} |S_i| = w \tag{11}$$

$$\wedge \bigwedge_{i=1}^{n-1} \bigwedge_{j=i+1}^{n} \left( |S_i \cap S_j| + |\overline{(S_i \cup S_j)}| \leq l - d \right) \wedge (S_i < S_j) \tag{12}$$

The constraints described by Equation (12) are implemented using a single ROBDD for each pair of $i$ and $j$ values. This is possible since we can model the integer addition and comparison operations as ROBDDs as described in Section 5, using the representation of the cardinality of a set variable as an integer expression as described in Section 5.5.

In order to find an optimal solution, we initially set $n$ to 1, and repeatedly solve instances of the problem, progressively incrementing $n$ to find larger and larger codes. We prove optimality of a solution for $n = k$ by failing to solve the problem for $n = k + 1$; we then know that the optimal value for $n$ was $k$.

Table 7: Statistics for the 51 Weighted Hamming Code testcases that were solved by all of the solvers.

| | Bounds | | Domain | | LU+R | | LU+Lex | | LU+Card | |
|---|---|---|---|---|---|---|---|---|---|---|
| | time /s | fails | time /s | fails | time /s | fails | time /s | fails | time /s | fails |
| Mean | 17.7 | 110034 | **0.2** | **210.7** | 0.3 | **210.7** | 4.4 | 1886 | 3.6 | 6604 |
| Total | 903.3 | | **11.4** | | 14.2 | | 222.6 | | 184.6 | |
| Minimum | **0.03** | 0 | **0.03** | 0 | **0.03** | 0 | **0.03** | 0 | **0.03** | 0 |
| 25th Percentile | **0.03** | 19 | **0.03** | 0 | **0.03** | 0 | **0.03** | 3 | **0.03** | 0 |
| Median | 0.05 | 196 | **0.04** | **2** | **0.04** | **2** | 0.04 | 20 | **0.04** | **2** |
| 75th Percentile | 0.67 | 5604 | **0.06** | **25** | **0.06** | **25** | 0.18 | 143.5 | 0.09 | 112 |
| Maximum | 415.55 | 3021057 | **4.16** | **3740** | 4.69 | **3740** | 113.9 | 45667 | 92.09 | 211677 |

Table 8: Weighted Hamming Code testcases that were solved by at least one but not all solvers

| | Bounds | | Domain | | LU+R | | LU+Lex | | LU+Card | |
|---|---|---|---|---|---|---|---|---|---|---|
| | time /s | fails | time /s | fails | time /s | fails | time /s | fails | time /s | fails |
| H(8,4,4) | — | — | **1.6** | **224** | 1.9 | **224** | 8.5 | 986 | — | — |
| H(9,4,3) | — | — | **11.3** | **5615** | 20.6 | **5615** | | | — | — |
| H(9,4,6) | — | — | **25.4** | **16554** | 45.7 | **16554** | 321.9 | 56599 | — | — |
| H(10,6,5) | — | — | **26.7** | **16635** | 29.6 | **16635** | 528.8 | 169457 | 428.3 | 762775 |

In order to assist with timing comparisons, we use the same set of problem instances as Sadler and Gervet (2004). We consider sets of values $H(l, d, w)$ where $l \in \{6, 7, 8, 9, 10\}$, $d \in \{4, 6, 8, 10, 12\}$, and $w \in \{3, 4, 5, 6, 7, 8\}$, with $d < l$ and $w \leq l$ (trivially there is at most one solution if $d \geq l$ and none if $w > l$). There are 62 such testcases; some of these are almost identical—in particular testcases $H(l, d, w)$ and $H(l, d, l - w)$ have solutions that differ only through complementation of the bits. The ROBDD-based solver can solve all but seven test cases (namely $H(9, 4, 4), H(9, 4, 5), H(10, 4, 3), H(10, 4, 4), H(10, 4, 5), H(10, 4, 6), H(10, 4, 7)$), which in reality contains three pairs of mirror image testcases.

Since there are too many results to list each testcase individually, performance statistics for the testcases that all the solvers were able to solve are shown in Table 7. Those cases that were solved by at least one but not all of the ROBDD-based solvers are shown in Table 8.

Sadler and Gervet (2004) report results for this problem using set bounds and lexicographic bounds solvers implemented in $ECL^iPS^e$. Their solvers were able to solve 50 of the testcases with a time limit of 240 seconds for each testcase. Some individual testcases took upwards of 100 seconds. By contrast, the ROBDD-based domain solver is capable of solving 55 testcases in 76.4 seconds in total. Clearly in this case enforcing domain consistency brings a considerable reduction in search space, leading to a highly efficient solver.

Moreover, the set bounds and lexicographic bounds solvers implemented using the ROBDD platform appear to perform considerably better than those of Sadler and Gervet. Due to the graphical nature of their presentation it is difficult to quantify this performance difference; exact figures are presented for only two testcases—namely $H(9,4,7)$ and $H(10,6,7)$. For the testcase $H(9,4,7)$ the ROBDD set bounds and lexicographic bounds solvers each found and proved an optimal solution in 0.6 and 0.3 seconds respectively, compared with the $> 240$ and 167.1 seconds respectively quoted for the bounds and lexicographic solvers of Sadler and Gervet. Similarly for $H(10,6,7)$ the ROBDD set bounds and lexicographic bounds solvers found and proved an optimal solution in 1.9 and 1.1 seconds respectively, as compared with $> 240$ and 98.5 seconds. Given this dramatic performance difference, it appears the additional modeling flexibility of the ROBDD-based solver provides a substantial performance gain.

It should be noted that the performance results reported by Sadler and Gervet were run on a slightly slower machine (a 2 Ghz Pentium 4 machine). Nonetheless, the contribution to the performance difference due to machine speed is dwarfed by the performance gap between the two solvers.

### 7.4 Balanced Academic Curricula

The Balanced Academic Curriculum problem (problem `prob030` of CSPLib) involves planning an academic curriculum over a sequence of academic periods in order to provide a balanced load in each period.

A curriculum consists of $m$ courses ($1 \leq i \leq m$) and $n$ academic periods. Each course $i$ has a set of prerequisites and an associated academic load $t_i$. Every course must be assigned to exactly one period. In any given period the total number of courses must be at least some minimum number $c$ and at most a maximum number $d$. In addition, within any given period the total academic load of the courses must be at least some minimum load $a$ and at most a maximum load $b$. Let $R$ be the set of prequisite pairs $\langle i,j \rangle$, where $i$ and $j$ are courses. Prerequisite relationships must be observed, so for each pair of courses $\langle i,j \rangle \in R$, if course $i$ is scheduled in period $p$, then course $j$ must be scheduled in a period strictly prior to $p$.

A model of this problem using set variables and set constraints was proposed by Hnich et al. (2002), although no experimental results were presented for that model. In this "primal" model, we use one set variable $S_i$ per academic period. Each $S_i$ represents the set of courses assigned to academic period $i$, so $k \in S_i$ if and only if course $k$ is assigned to period $i$. We can model the problem using the following constraints:

- (S1) Every course is taken exactly once: $\texttt{partition}(S_1, \ldots, S_n)$

- (S2) The number of courses in a period is between $c$ and $d$:
  $\forall_{i=1}^{n}((c \leq |S_i|) \land (|S_i| \leq d))$

- (S3) The total academic load in a period is between $a$ and $b$:
  $\forall_{i=1}^{n}((a \leq \texttt{wsum}(S_i, \langle t_1, \ldots, t_m \rangle)) \land (\texttt{wsum}(S_i, \langle t_1, \ldots, t_m \rangle) \leq b))$

- (S4) Prerequisites are respected:
  $\forall_{i=1}^{n}\forall_{j=1}^{i}(\bigwedge_{\langle c,p \rangle \in R}(p \in S_i) \to (c \notin S_j))$

In general the `partition` and `wsum` constraints of the primal set model can be very large, making domain propagation impractical.

In addition to presenting results for the primal model, Choi et al. (2003) demonstrated that a substantial performance improvement can be obtained for this problem through the use of redundant models together with channelling constraints.

We can obtain better results from a "dual" set model, where additional set variables are introduced to model each course. We define set variables $X_i$ ($1 \leq i \leq m$) representing each course, such that if $k \in X_i$ then course $i$ is assigned to period $k$. We can then define the following constraints:

- (CX) Channelling constraints: $\forall_{i=1}^m \forall_{j=1}^n (i \in S_j) \leftrightarrow (j \in X_i)$

- (X1) Each course may be assigned to at most one period: $\forall_{i=1}^m |X_i| = 1$

- (X2) Prerequisites are respected: $\forall \langle i, j \rangle \in R$ $lexlt(X_j, X_i)$

We define the dual model to consist of the constraints {S2, S3, CX, X1, X2}. Constraints S1 and S4 are no longer required, since they are propagation redundant.

Unfortunately, while the dual model performs better than the primal set model, it is still incapable of proving the optimality of the solutions to any of the problems. We can obtain stronger propagation by introducing redundant global constraints on the total academic load and number of courses of all academic periods, as suggested by Choi et al. (2003).

For each academic period $j$ define two integer variables $l_j$, representing the academic load in period $j$ and $q_j$, representing the number of courses in period $j$. We can then define the following constraints on these new integer variables:

- (CI1) Channeling constraints on $l_i$: $\forall_{i=1}^n \mathtt{wsum}(S_i, \langle t_1, \ldots, t_m \rangle) = l_i$

- (CI2) Channeling constraints on $q_i$: $\forall_{i=1}^n |S_i| = q_i$

- (I1) Range constraints on $l_i$: $\forall_{i=1}^n (a \leq l_i) \wedge (l_i \leq b)$

- (I2) Range constraints on $q_i$: $\forall_{i=1}^n (c \leq q_i) \wedge (q_i \leq d)$

- (I3) All loads should be undertaken: $\sum_{i=1}^n l_j = \sum_{i=1}^n t_i$

- (I4) All courses must be taken: $\sum_{i=1}^n q_j = m$

We then define the "hybrid dual" model consisting of the constraints {CX, X1, X2, CI1, CI2, I1, I2, I3, I4}. Note that none of the original constraints S1–S4 remain. For completeness it is also possible to define a "hybrid primal" model consisting of constraints {S1, S4, CI1, CI2, I1, I2, I3, I4}, although as before the large ROBDD sizes make domain propagation impractical.

Experimental results for the Balanced Academic Curriculum problem are shown in Table 9. The timing results for the Hybrid Dual model are comparable to the best results obtained by Hnich et al. (2002) and Choi et al. (2003). The number of failures required to solve the problem is orders of magnitude smaller than the best results presented in either paper, emphasising the value of domain propagation in this case. These examples also show a case where the LU+Lex solver is competitive and clearly the most robust solver over the different models.

Table 9: Performance results for the Balanced Academic Curriculum Problem for the 8, 10 and 12 period problems. The first column contains the number of periods, the second column contains the model type (HD=hybrid dual, D=dual, P=primal, HP=hybrid primal), and the third column contains the maximum load per period *b*. In all cases a sequential "smallest-element-in-set" labeling method was used

| Problem | | | Bounds | | Domain | | LU+R | LU+Lex | | LU+Card | |
|---|---|---|---|---|---|---|---|---|---|---|---|
| | | | time /s | fails | time /s | fails | time /s | time /s | fails | time /s | fails |
| **8** | HD | 16 | — | — | **0.3** | **0** | **0.3** | **0.3** | **0** | — | — |
| | | 17 | 8.1 | 5586 | 1.7 | **3** | 2.7 | **1.0** | **3** | 2.3 | **3** |
| | | 18 | 35.7 | 30430 | 1.7 | **5** | 3.0 | **1.0** | **5** | 2.3 | **5** |
| | D | 16 | — | — | — | — | — | — | — | — | — |
| | | 17 | 7.7 | 5348 | 1.1 | **3** | 1.2 | **0.6** | **3** | 1.7 | **3** |
| | | 18 | 32.8 | 29781 | 1.2 | **5** | 1.3 | **0.7** | **5** | 1.6 | **5** |
| | P | 16 | — | — | — | — | — | — | — | — | — |
| | | 17 | **6.0** | 5353 | — | — | — | 13.9 | **2046** | 32.7 | 4757 |
| | | 18 | **16.3** | 29781 | — | — | — | 53.1 | **26431** | 61.2 | 29628 |
| | HP | 16 | — | — | — | — | — | **0.3** | **0** | — | — |
| | | 17 | **6.8** | 5586 | — | — | — | 12.8 | **1510** | 40.5 | 5410 |
| | | 18 | **19.3** | 30431 | — | — | — | 69.8 | **26145** | 70.5 | 30141 |
| **10** | HD | 13 | — | — | **0.3** | **0** | **0.3** | **0.3** | **0** | — | — |
| | | 14 | 17.2 | 11439 | 1.3 | **2** | 1.5 | **0.9** | **2** | 2.6 | **2** |
| | | 15 | 2.1 | 630 | 1.4 | **1** | 1.7 | **0.9** | **1** | 2.6 | **1** |
| | D | 13 | — | — | — | — | — | — | — | — | — |
| | | 14 | 26.8 | 21103 | 1.0 | **2** | 1.6 | **0.6** | **2** | 2.4 | **2** |
| | | 15 | 2.1 | 711 | 1.0 | **1** | 1.6 | **0.6** | **1** | 2.1 | **1** |
| | P | 13 | — | — | — | — | — | — | — | — | — |
| | | 14 | **14.7** | 21105 | — | — | — | 53.3 | 16275 | 52.1 | **11772** |
| | | 15 | **3.5** | 711 | — | — | — | 9.6 | 593 | 31.5 | **515** |
| | HP | 13 | — | — | — | — | — | **0.4** | **0** | — | — |
| | | 14 | **11.8** | 11440 | — | — | — | 25.6 | **4672** | 46.9 | 5773 |
| | | 15 | **3.6** | 630 | — | — | — | 10.6 | **516** | 31.7 | 533 |
| **12** | HD | 16 | — | — | 1.1 | **0** | 1.1 | **0.9** | **0** | — | — |
| | | 17 | — | — | 6.0 | **9** | 9.3 | **3.0** | **9** | — | — |
| | | 18 | 6.1 | 235 | 8.1 | **4** | 9.2 | **3.4** | **4** | 15.7 | **4** |
| | D | 16 | — | — | — | — | — | — | — | — | — |
| | | 17 | — | — | — | — | — | — | — | — | — |
| | | 18 | 6.3 | 220 | 6.8 | **4** | 9.0 | **3.0** | **4** | 13.1 | **4** |
| | P | 16 | — | — | — | — | — | — | — | — | — |
| | | 17 | — | — | — | — | — | — | — | — | — |
| | | 18 | **101.7** | 225 | — | — | — | 153.8 | **224** | — | — |
| | HP | 16 | — | — | — | — | — | **3.4** | **0** | — | — |
| | | 17 | — | — | — | — | — | **216.8** | **526** | — | — |
| | | 18 | **100.9** | 236 | — | — | — | 166.0 | **224** | — | — |

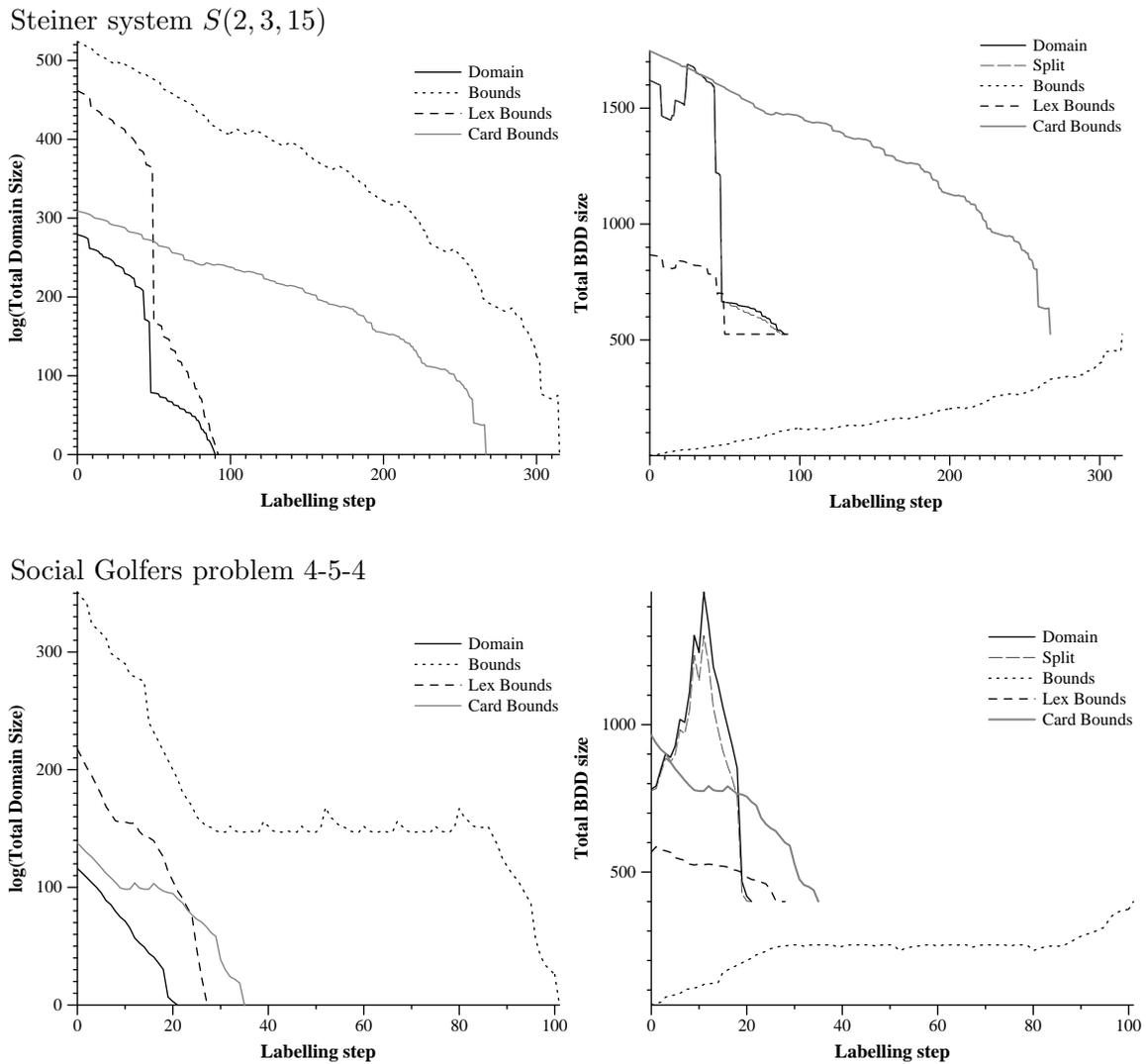

Figure 17: Comparison of total domain and total ROBDD sizes with labeling step for set bounds, set domain, split set domain, set and lexicographic bounds, and set and cardinality bounds solvers on the Steiner System $S(2,3,15)$ and Social Golfers problem 4-5-4. Note that the $y$ axes of the total domain size graphs have a logarithmic scale (base 2).

## 7.5 Comparing the Propagation Performance of the Solvers

It is instructive to compare the propagation performance of the various ROBDD-based solvers graphically. Here we pick two small examples, namely the Steiner System $S(2,3,15)$ and the Social Golfers problem 4-5-4, using the default labeling for each problem, and graph BDD and domain sizes against number of labeling steps and time.

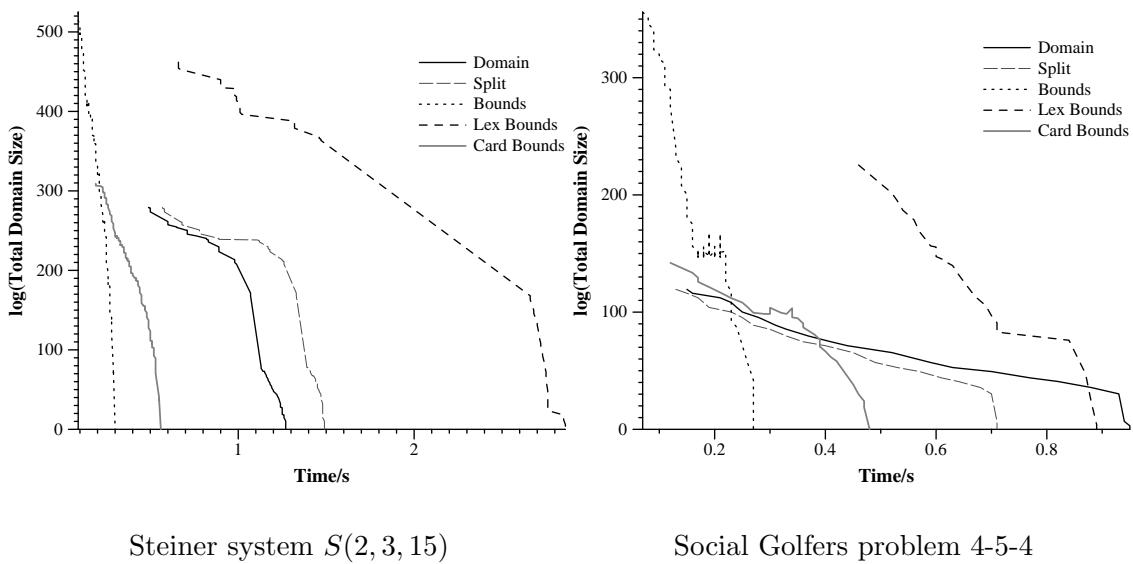

Steiner system $S(2, 3, 15)$          Social Golfers problem 4-5-4

Figure 18:  Comparison of total domain size with time for set bounds, set domain, split set domain, set and lexicographic bounds, and set and cardinality bounds solvers on the Steiner System $S(2, 3, 15)$ and Social Golfers problem 4-5-4. Note that the $y$ axes have a logarithmic scale (base 2).

Figure 17 depicts how the total domain sizes and total ROBDD sizes (total number of internal nodes) vary with each labeling step for each of the solvers for the Steiner System $S(2, 3, 15)$ and for the Social Golfers problem 4-5-4. In particular, observe that the domain and split-domain propagators are the most effective at reducing the domain size and thus restricting the search space, although maintaining domain consistency can be costly due to the size of the ROBDDs required for representing arbitrary domains. This can be seen in the ROBDD size against labeling step graph, most notably in the case of the Social Golfers problem 4-5-4. The lexicographic bounds solver is the next most effective in domain size reduction, and requires a smaller number of ROBDD nodes to store its bounds than the domain or cardinality bounds solvers. The bounds solver is clearly the weakest in terms of domain size reduction, and as opposed to the other solvers starts from a zero size domain representation and builds slowly to ROBDDs representing the answer.

We are not only concerned with the strength of a propagator in the abstract, since the efficiency of the solver as a whole is dependent on both the propagation and labeling processes. Figure 18 depicts graphs of domain size against time for each of the propagators discussed in this paper. Here we see that the most effective propagator does not necessarily lead to the most efficient set solver. Despite comparatively weak propagation, the set and cardinality bounds solvers lead to the fastest reduction in domain size per unit time. Nonetheless, the experimental timing results given above demonstrate that for harder cases the additional cost of maintaining domain consistency or lexicographic bounds can be justified through the consequent reduction in search space.

Interestingly we can see from the graphs in Figure 18 that the initial domain reduction before labeling, which is the gap from time 0 to where the line starts, can be a significant part of the computation. In particular, one of the weaknesses of the lexicographic bounds solver is the large time required to reach an initial fixpoint through lexicographic bounds propagation. For practical applications it might be preferable to implement a hybrid solver which utilises one of the other propagators to generate the initial domains, and uses the lexicographic bounds propagator during labeling.

## 8. Conclusion

We have demonstrated that ROBDDs form a highly flexible platform for constructing set constraint solvers. ROBDDs allow a compact representation of many set domains and set constraints, making them an effective basis for an efficient set solver. Since ROBDDs can represent arbitrary Boolean formulæ we can easily conjoin and existentially quantify them, permitting the removal of intermediate variables and making the construction of global constraints trivial. We have demonstrated how to efficiently enforce various levels of consistency, including set domain, set bounds, cardinality bounds and lexicographic bounds consistency. Finally, we have presented experimental results that demonstrate that the ROBDD-based solver outperforms other common set solvers on a wide variety of standard set constraint satisfaction problems.

No single set solver is uniformly better than the others. For many examples simple bounds propagation is the best approach, while in other cases, particularly when we ask for all solutions, domain consistency is preferable. There are also examples where lexicographic bounds or cardinality bounds are the best approach. The split-domain approach is somewhat disappointing, since it appears that in many cases the overhead of the more complicated calculation (Equation 6) is not rewarded in terms of smaller ROBDD sizes.

In the future we plan to explore a robust general set constraint solver that dynamically chooses which level of consistency to maintain by examining how big the domain ROBDDs are becoming as search progresses.

## References

Andersen, H. R. (1998). An introduction to Binary Decision Diagrams. [Online, accessed 30 July 2004]. `http://www.itu.dk/people/hra/notes-index.html`.

Apt, K. R. (1999). The essence of constraint propagation. *Theoretical Computer Science*, *221*(1–2), 179–210.

Azevedo, F. (2002). *Constraint Solving over Multi-valued Logics*. Ph.D. thesis, Faculdade de Ciências e Tecnologia, Universidade Nova de Lisboa.

Bagnara, R. (1996). A reactive implementation of Pos using ROBDDs. In *Procs. of PLILP*, Vol. 1140 of *LNCS*, pp. 107–121. Springer.

Bessiere, C., Hebrard, E., Hnich, B., & Walsh, T. (2004). Disjoint, partition and intersection constraints for set and multiset variables. In Wallace, M. (Ed.), *Proceedings of the*

*10th International Conference on Principles and Practice of Constraint Programming*, Vol. 3258 of *LNCS*, pp. 138–152. Springer-Verlag.

Bryant, R. E. (1986). Graph-based algorithms for Boolean function manipulation. *IEEE Trans. Comput.*, *35*(8), 677–691.

Bryant, R. E. (1992). Symbolic Boolean manipulation with ordered binary-decision diagrams. *ACM Comput. Surv.*, *24*(3), 293–318.

Choi, C., Lee, J., & Stuckey, P. J. (2003). Propagation redundancy in redundant modelling. In Rossi, F. (Ed.), *Proceedings of the 9th International Conference on Principles and Practices of Constraint Programming*, Vol. 2833 of *LNCS*, pp. 229–243. Springer-Verlag.

Gent, I. P., Walsh, T., & Selman, B. (2004). CSPLib: a problem library for constraints. [Online, accessed 24 Jul 2004]. `http://www.csplib.org/`.

Gervet, C. (1997). Interval propagation to reason about sets: Definition and implementation of a practical language. *Constraints*, *1*(3), 191–246.

Hawkins, P., Lagoon, V., & Stuckey, P. (2004). Set bounds and (split) set domain propagation using ROBDDs. In Webb, G., & Yu, X. (Eds.), *AI 2004: Advances in Artificial Intelligence, 17th Australian Joint Conference on Artificial Intelligence*, Vol. 3339 of *LNCS*, pp. 706–717. Springer-Verlag.

Hnich, B., Kiziltan, Z., & Walsh, T. (2002). Modelling a balanced academic curriculum problem. In *Proceedings of the Fourth International Workshop on Integration of AI and OR Techniques in Constraint Programming for Combinatorial Optimization Problems*, pp. 121–131.

IC-PARC (2003). The ECLiPSe constraint logic programming system. [Online, accessed 31 May 2004]. `http://www.icparc.ic.ac.uk/eclipse/`.

ILOG (2004). ILOG Solver. [Online, accessed 30 Aug 2004]. `http://www.ilog.com/`.

Kiziltan, Z., & Walsh, T. (2002). Constraint programming with multisets. In *Proceedings of the 2nd International Workshop on Symmetry in Constraint Satisfaction Problems (SymCon-02)*.

Lagoon, V., & Stuckey, P. (2004). Set domain propagation using ROBDDs. In Wallace, M. (Ed.), *Proceedings of the 10th International Conference on Principles and Practice of Constraint Programming*, Vol. 3258 of *LNCS*, pp. 347–361. Springer-Verlag.

van Lint, J. H., & Wilson, R. M. (2001). *A Course in Combinatorics* (2nd edition). Cambridge University Press.

Mailharro, D. (1998). A classification and constraint-based framework for configuration. *Artificial Intelligence for Engineering Design, Analysis and Manufacturing*, *12*, 383–397.

Müller, T. (2001). *Constraint Propagation in Mozart*. Doctoral dissertation, Universität des Saarlandes, Naturwissenschaftlich-Technische Fakultät I, Fachrichtung Informatik, Saarbrücken, Germany.

Müller, T., & Müller, M. (1997). Finite set constraints in Oz. In Bry, F., Freitag, B., & Seipel, D. (Eds.), *Workshop Logische Programmierung*, Vol. 13. Technische Universität München.

Puget, J.-F. (1992). PECOS: a high level constraint programming language. In *Proceedings of SPICIS'92*, Singapore.

Sadler, A., & Gervet, C. (2001). Global reasoning on sets. In *FORMUL'01 workshop on modelling and problem formulation, in conjunction with CP'01*.

Sadler, A., & Gervet, C. (2004). Hybrid set domains to strengthen constraint propagation and reduce symmetries. In Wallace, M. (Ed.), *Proceedings of the 10th International Conference on Principles and Practice of Constraint Programming*, Vol. 3258 of *LNCS*, pp. 604–618. Springer-Verlag.

Somenzi, F. (2004). CUDD: Colorado University Decision Diagram package. [Online, accessed 31 May 2004]. `http://vlsi.colorado.edu/~fabio/CUDD/`.

Somogyi, Z., Henderson, F., & Conway, T. (1996). The execution algorithm of Mercury, an efficient purely declarative logic programming language. *Journal of Logic Programming*, *29*(1–3), 17–64.